%% file: main.tex
\documentclass[11pt]{article}
\usepackage[final]{acl}

\usepackage{times}
\usepackage{latexsym}
\usepackage[T1]{fontenc}
\usepackage[utf8]{inputenc}
\usepackage{microtype}
\usepackage{graphicx}
\usepackage{booktabs}
\usepackage{amsmath}
\usepackage{amssymb}
\usepackage{xcolor}
\usepackage{url}
\usepackage{enumitem}
\usepackage{tikz}
\usetikzlibrary{arrows.meta,positioning,calc}

\hypersetup{
  pdftitle={Peak-Then-Collapse and the Four Interface Channels of Knowledge-Graph Tool Use},
  pdfauthor={Tianda Sun and Dimitar Kazakov}
}

\newcommand{\thinktag}{\itshape\color{gray!60!black}}
\newcommand{\searchtag}{\ttfamily\color{blue!55!black}}
\newcommand{\responsetag}{\ttfamily\color{teal!55!black}}
\newcommand{\answertag}{\ttfamily\bfseries\color{red!60!black}}

\renewcommand{\arraystretch}{0.95}
\setlist[itemize]{itemsep=2pt,topsep=2pt,parsep=0pt}

\title{Peak-Then-Collapse and the Four Interface Channels of\\
Knowledge-Graph Tool Use}

\author{
  Tianda Sun \\
  University of York \\
  Department of Computer Science \\
  Heslington, York \\
  YO10 5DD \\
  \texttt{tianda.sun@york.ac.uk} \\\And
  Dimitar Kazakov \\
  University of York \\
  Department of Computer Science \\
  Heslington, York \\
  YO10 5DD \\
  \texttt{dimitar.kazakov@york.ac.uk} \\
}

\begin{document}
\maketitle

\begin{abstract}
We test the standard RLVR tool-use recipe (GRPO on Qwen2.5-7B-Instruct) on a deliberately minimal knowledge-graph tool API: four Freebase navigation verbs over Complex WebQuestions. Under a self-verifiable retrieval reward (R-selfV), the policy's correct-via-tool rate climbs from $3.8\%$ to $9.6\%$ over 250 steps, then collapses to $0\%$ within a single 50-step window, a \emph{peak-then-collapse} pattern replicated across four seeds. Over the same window, strict EM falls $39.5{\to}0.0\%$ and containment EM falls $46.1{\to}12.1\%$. Across seven reward designs, we find four recurring failure modes: adding denser or more targeted proxy rewards shifts the failure mode rather than eliminating it. We argue that a key difference from Python interpreters, web search, and JSON APIs is interface feedback: their failures often leak natural-language signal the model saw in pretraining. A Python traceback names the failing line; an empty Freebase result \texttt{[]} does not. Within the seven-design ladder, stripping away that surface exposes a degradation regime not removed by the same-family redesigns. Full-test relation- and entity-replacement oracles change EM by only $+0.20$ and $+0.42$~pp, and $95.4\%$ of retrieval-dependent errors are retrieval-composition rather than answer-extraction failures. As mitigation, one-iteration self-distillation reaches $40.0\%$ EM at 7B; a differently configured 14B run reaches $40.2\%$, so this comparison does not isolate capacity.
\end{abstract}

\input{sections/01_introduction}
\input{sections/02_related_work}
\input{sections/03_methodology}
\input{sections/04_main_results}
\input{sections/06_framework}
\input{sections/10_conclusion}

\section*{Limitations}
\input{sections/09_limitations}

\section*{Acknowledgments}
We thank the reviewers and Area Chair for their constructive feedback.
In preparing this work, the authors used AI-based assistants for language editing of the manuscript (drafting and revising prose and \LaTeX{} formatting) and for coding assistance (implementing and debugging the training and evaluation pipelines), under the authors' direction. All research questions, experimental design, analyses, and claims are the authors' own; the authors verified all AI-assisted text and code for correctness and take full responsibility for the content of this paper.

\bibliography{custom}

\appendix
\input{appendix/appendix_B_searchr1_options}
\input{appendix/appendix_C_rewards_and_classifier}
\input{appendix/appendix_D_sft_corpus}
\input{appendix/appendix_F_gpt4o_baseline}
\input{appendix/appendix_G_framework_illustration}
\input{appendix/appendix_I_llama_cross_family}
\input{appendix/appendix_H_trajectory_exemplars}

\end{document}

%% file: sections/01_introduction.tex
\section{Introduction}
\label{sec:intro}

Large language models increasingly act as \emph{agents}: they call external tools (code interpreters, web search, function APIs) and are trained to use them by reinforcement learning with verifiable rewards (RLVR), where a checkable outcome (did the code run? did the answer match?) supplies the reward~\citep{li2025torl,feng2025retool,jin2025searchr1,qian2025toolrl}. The dominant recipe (dense process rewards, GRPO~\citep{shao2024grpo}, a 7B-scale policy) is treated as portable across tool families. We test that portability on a knowledge graph (KG), whose interface is by construction the opposite of the pretraining-aligned surfaces RLVR usually enjoys. Our benchmark is Complex WebQuestions (CWQ)~\citep{talmor2018cwq}, a standard multi-hop question-answering task over the Freebase~\citep{bollacker2008freebase} knowledge graph, which we re-cast from its usual gold-subgraph/SPARQL form into a \emph{deliberately minimal tool-use interface}: the policy must reach answers by composing four navigation verbs (the outgoing or incoming relations of an entity, and the entities reachable from it along a relation), where every entity is an opaque machine ID (\texttt{m.0d3k14}) and every miss returns an uninformative empty list. Python tool use leaks natural-language signal the model saw in pretraining (a traceback names the failing line), whereas this KG interface leaks almost none. We ask whether the standard RLVR-tool recipe still converges once that surface is removed.

\begin{figure}[t]
\centering
\includegraphics[width=0.82\columnwidth]{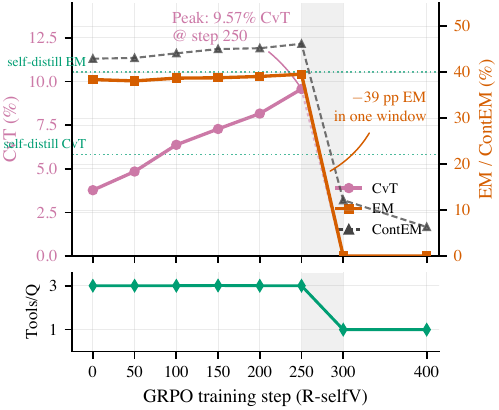}
\caption{\textbf{Peak-then-collapse for Qwen2.5-7B under R-selfV.} Top: CvT (correct-via-tool rate; left, blue) climbs $3.77{\to}9.57\%$ (steps 0--250) then falls to $0$ by step~300; strict EM falls $39.51{\to}0.00\%$ while ContEM falls $46.13{\to}12.14\%$. Bottom: Tools/Q drops $3.0{\to}1.0$ across the shaded 50-step window. self-distill (dashed) reaches $40.0\%$ EM / $5.81\%$ CvT in the same reward family. Mechanism: \S\ref{sec:taxonomy-mode4}.}
\label{fig:collapse}
\end{figure}

Across seven reward designs trained on Qwen2.5-7B-Instruct~\citep{qwen2025qwen25} + CWQ, we find four distinct failure modes, none solved by the next reward design in the same family. The signature finding is a Qwen/R-selfV \emph{peak-then-collapse} (\S\ref{sec:taxonomy-mode4}): a self-verifiable retrieval reward (one that checks whether a returned entity appears verbatim in the answer, gold-free) lifts \emph{grounded retrieval} (the correct-via-tool rate, CvT) from $3.8\%$ to $9.6\%$ over 250 training steps, then collapses to $0\%$ in a single 50-step window across four seeds. In that window, strict EM falls $39.5{\to}0.0\%$ and containment EM (ContEM) falls $46.1{\to}12.1\%$ (Figure~\ref{fig:collapse}).

Agentic-RL collapse phenomena have been documented~\citep{ragen2025,jin2025searchr1,ragen2_2026}, but at the level of action-distribution entropy and reward variance. The specific mechanism we observe, a one-shot quote-and-stop attractor with mean tool calls per question (Tools/Q) dropping $3.0{\to}1.0$ in a 50-step window, is a call-volume signature distinct from these precedents and from the standard RLHF reward-hacking framework~\citep{gao2023scaling,pan2022effects,skalse2022defining}, which reports proxy-overoptimisation at the level of learned reward models. Across the seven-design ladder, stripping away the pretraining-aligned surface exposes a degradation regime not removed by the same-family redesigns in the main experiment.

We run a controlled comparison: one primary model, one task, one algorithm (GRPO), one tool API (the 4-verb Freebase interface), and seven reward designs read as a pragmatic ladder (\S\ref{sec:setup-ladder}; two rungs also vary the KL tether or initialisation). To explain the convergent failures, we formalise the tool-API-to-policy-gradient pathway as four channels along which a tool API supplies (or fails to supply) RL signal: silent failure, symbolic schema, opaque compositional state, and absent pretraining prior ($L_{\mathrm{sig/lang/comp/prior}}$). We use the channels to organise the observed failure surfaces (\S\ref{sec:framework}). At the strongest in-family checkpoint, replacing relation arguments with gold relations changes EM by only $+0.20$~pp; a complementary entity-replacement oracle changes it by $+0.42$~pp. Together with the error decomposition, these results point to joint retrieval-composition failures.

\paragraph{Contributions.}
\textbf{(C1)} A controlled comparison of seven reward designs on a single primary model $\times$ task $\times$ algorithm $\times$ tool API, organised into a four-mode failure taxonomy (\S\ref{sec:taxonomy}).
\textbf{(C2)} A signal-theoretic four-channel framework supported by relation- and entity-replacement oracles, plus a step-level error decomposition where $95.4\%$ of $1{,}958$ \emph{retrieval-dependent} errors localise to retrieval composition rather than answer extraction (\S\ref{sec:framework-evidence}).
\textbf{(C3)} A one-iteration ReST-EM-style self-distillation recipe reaching EM=$40.0\%$ at 7B; the 5K rule-based pre-SFT prefix is dispensable. A differently configured 14B run reaches $40.2\%$ (\S\ref{sec:g2}).
\textbf{(C4)} Step-resolved (50-step cadence, full CWQ test set) Qwen/R-selfV peak-then-collapse across four seeds: a call-volume signature (Tools/Q $3.0{\to}1.0$) novel relative to the action-distribution entropy and reward-variance diagnostics of prior agentic-RL collapse~\citep{ragen2025,jin2025searchr1,ragen2_2026} (\S\ref{sec:taxonomy-mode4}).

We release all code, reward implementations, per-checkpoint evaluations, and the seven-category trajectory classifier at \url{https://github.com/TiandaSun/KG_GRPO}.

%% file: sections/02_related_work.tex
\section{Related Work}
\label{sec:related}

\paragraph{Process-reward RL for tools and knowledge graphs.}
Process-reward recipes for tool-using LLMs~\citep{jin2025searchr1,chen2025research,deepseek2025r1,wang2025stepsearch,qian2025toolrl,zhang2025criticsearch} succeed on tool APIs with pretraining-aligned surfaces; we test whether the same recipe transfers to KG tools deliberately stripped of natural-language surface. KG-specific variants using KGs to \emph{define} rewards (fs1~\citep{zhang2025followpath}; HyperGraphPro~\citep{park2026hypergraphpro}; KG-Implicit-RM~\citep{kansal2026kgimplicitrm}) differ from our setup, which uses the KG to evaluate the policy (under RoG's~\citep{luo2024rog} subgraph filter; strict EM motivated by~\citet{zhang2025pitfalls}). On the same Qwen2.5-7B-Instruct + CWQ benchmark, richer-pipeline concurrent systems report higher numbers while changing several components (GraphWalker~\citep{graphwalker2026} adds $\sim 21$K-trajectory SFT + BM25 + an LLM reranker, $79.6\%$ EM; Graph-RFT~\citep{graphrft2025} adds web fallback over a $40\%$-incomplete KG, $67.2\%$ Hits@1). These cross-system comparisons motivate studying interface enrichment, but do not isolate it from data, training, or evaluation differences. Our stripped-interface diagnostic is a deliberate complement.

\paragraph{Cold-start SFT-then-RL.}
Our self-distillation recipe sits in the cold-start-SFT-then-RL family~\citep{feng2025retool,li2025torl,chen2023fireact}; the methodological extension is that our SFT corpus is \emph{self-distilled} from our own RL checkpoint with a strict EM-correct $\cap$ tool-productive $\cap$ format-valid filter, connecting to ReST-EM~\citep{singh2024restem}. The no-rule-prefix variant (\S\ref{sec:g2}) shows that removing the rule-SFT prefix while retaining self-distillation SFT reaches the same approximately $40\%$ EM band, consistent with ReST-EM's init-from-base prescription on a different task family.

\paragraph{Reward hacking and agentic-RL collapse.}
The peak-then-collapse mode (\S\ref{sec:taxonomy-mode4}) extends a known reward-hacking pattern. RLHF~\citep{skalse2022defining,pan2022effects,gao2023scaling} catalogues proxy-overoptimisation at the learned-reward-model level; agentic-RL diagnostics include action-distribution entropy~\citep{yu2025demystifying}, reward variance (Echo Trap~\citealp{ragen2025}; \citealp{ragen2_2026,deng2025llds}), and metric-without-capability gain~\citep{shao2025spurious} (closely related to our R-stepwise tool-free shortcut, \S\ref{sec:taxonomy-mode2}); \citet{yue2025rlcapacity} (RL on math does not push pass@$k$ beyond base) is our closest non-KG analogue. Our peak-then-collapse contribution is the Tools/Q call-volume signature ($3.0{\to}1.0$) and step-resolved cadence on a KG-tool agent.

%% file: sections/03_methodology.tex
\section{Methodology}
\label{sec:methodology}

\subsection{Task and KG tool interface}
\label{sec:setup-task}

We use Complex WebQuestions (CWQ)~\citep{talmor2018cwq} over Freebase~\citep{bollacker2008freebase}, following the standard RoG~\citep{luo2024rog} splits ($27{,}639$ train / $3{,}531$ test).\footnote{$27{,}639$ is RoG's row-expanded count over (question, gold-path) pairs; the canonical \texttt{rmanluo/RoG-cwq} release reports $16{,}894$ unique training questions. Talmor \& Berant's original v1.1 release uses $27{,}734$ / $3{,}480$ / $3{,}475$; the residual differences are Freebase-reachability filtering.} Answers are normalised (lowercase, strip articles, strip punctuation, whitespace-collapse) before exact-match scoring. The RoG subgraph filter yields $2.59$M entities, $7{,}058$ relations, $8.3$M triples accessible at inference time.

The model accesses the KG through a \emph{four-verb} tool interface:
\begin{itemize}[leftmargin=*,itemsep=1pt,topsep=2pt]
\item \texttt{get\_tail\_relations(e)}, \texttt{get\_head\_relations(e)}: list relations outgoing from / incoming into entity $e$.
\item \texttt{get\_tail\_entities(e, r)}, \texttt{get\_head\_entities(e, r)}: list entities reachable from $e$ along relation $r$ in either direction.
\end{itemize}
Tool calls are emitted as \texttt{<search>\textit{verb}(\textit{args})</search>} tokens; the KG response is inlined verbatim into the context and generation continues. Up to five tool-calling turns are permitted per question. The interface is deliberately minimal: four verbs, one symbol per query, responses that list entities or relations line-by-line. Unlike a web-search RAG surface, every query response can be objectively classified as successful, empty, or off-schema, the property that makes step-resolved reward analysis tractable. Figure~\ref{fig:setup} (App.~\ref{app:framework-illustration}) illustrates the re-cast.

\subsection{Shared SFT initialisation}
\label{sec:setup-sft}

All reward-family ablations share a single SFT-initialised Qwen2.5-7B-Instruct~\citep{qwen2025qwen25} checkpoint with LoRA ($r{=}64$, $\alpha{=}128$, DoRA + rslora) on $5{,}000$ rule-based trajectories whose \texttt{<search>} calls are constructed \emph{from the gold triple chain} (no LLM teacher: no GPT-4, no Claude, no Qwen-72B distillation). The rule-based generator and the $14{,}082$-trajectory G-variant corpus are in Appendix~\ref{app:sft}. A Qwen2.5-14B configuration (Table~\ref{tab:main}) and a Llama-3.1-8B-Base boundary check (App.~\ref{app:llama-cross-family}) use model-specific SFT pre-warms and different reward protocols, probing configuration and model-family boundaries around the primary Qwen/R-selfV setting.

\subsection{GRPO training}
\label{sec:setup-grpo}

GRPO~\citep{shao2024grpo} on the SFT base: group size $8$, \texttt{low\_var\_kl} loss, KL coefficient $0.05$ unless noted (Table~\ref{tab:main} for per-variant overrides), lr $3\mathrm{e}{-7}$, batch $128$, $500$ steps; \texttt{verl}~\citep{sheng2024verl} + vLLM~\citep{kwon2023vllm} rollouts on NVIDIA GH200. Reward implementations are unit-tested ($18/18$ passing; Appendix~\ref{app:rewards}).

\subsection{Reward ladder}
\label{sec:setup-ladder}

Intuitively, the ladder first tests whether final-answer reward alone preserves the tool format, then adds verifiable tool-call structure, then rewards productive tool use, then stabilises the policy with a stronger KL tether. R-selfV tests whether a gold-free retrieved-entity signal can improve grounding, while self-distill tests whether successful trajectories can stabilise the best in-family policy. Table~\ref{tab:ladder} summarises the six-rung design.

\begin{table*}[t]
\centering
\footnotesize
\setlength{\tabcolsep}{8pt}
\begin{tabular}{@{}clp{5.5cm}ccl@{}}
\toprule
Rung & Name & Patch over predecessor & EM & CvT & Mode emerged \\
\midrule
1 & R-binary                       & $r_\mathrm{out}$ only                                & $0.000$ & ---     & Mode 1 (format)  \\
2 & R-stepwise                     & $+\,r_\mathrm{valid}+r_\mathrm{path}+r_\mathrm{coh}$ & $0.325$ & $0.03\%$  & Mode 2 (ritual)  \\
3 & R-toolverbs                    & $+\,r_\mathrm{tool\text{-}type}$ bonus               & $0.322$ & $3.03\%$  & Mode 3 (drift)   \\
4 & R-toolverbs·KL                 & KL $\times 5$ (SFT-prior tether)                     & $0.384$ & $3.77\%$  & stable to $400+$ \\
5 & R-selfV$^\star$                & $+\,r_\mathrm{retrv}$ (entity in answer)             & $0.395$ & $\mathbf{9.57\%}$  & Mode 4 (cliff)   \\
6 & \textbf{self-distill}$^\star$  & R-toolverbs + self-distilled SFT pre-warm            & $\mathbf{0.400}$ & $5.81\%$  & stable (best EM) \\
\bottomrule
\end{tabular}
\caption{\textbf{Reward-ladder design.} Each rung patches a named failure of its predecessor (\S\ref{sec:taxonomy}). $^\star$Rungs 5--6 fork from R-toolverbs·KL@400 (R-selfV initialises from it; self-distill self-distills from its rollouts then re-inits from base). EM/CvT in Table~\ref{tab:main}; formulas in App.~\ref{app:rewards}.}
\label{tab:ladder}
\end{table*}

We organise reward design as a six-rung ladder (Table~\ref{tab:ladder}); each rung is introduced as a patch for a specific named failure of its predecessor (failure modes in \S\ref{sec:taxonomy}; full reward formulas in App.~\ref{app:rewards}). Rungs~4 and~6 also vary non-reward ingredients (KL coefficient; initialisation/data regime), so we read the ladder \emph{pragmatically} (does each patch fix its predecessor's failure?) rather than as strictly single-variable reward ablations.\footnote{Three earlier variants (E2 heuristic, E4 random-reward control, E5a retrieval-grounded cold-start) were run but cut as redundant; see App.~\ref{app:cut-variants}.}
\textbf{(1) R-binary}: outcome-only $r{=}0.5\,\mathrm{EM}{+}0.5\,\mathrm{F1}$ (Search-R1~\citep{jin2025searchr1}, DeepSeek-R1-zero~\citep{deepseek2025r1} family).
\textbf{(2) R-stepwise}: equal-weight $r_\mathrm{out}{+}r_\mathrm{valid}{+}r_\mathrm{path}{+}r_\mathrm{coh}$ ($r_\mathrm{valid}$ checks call syntax, $r_\mathrm{path}$ rewards gold-relation queries, $r_\mathrm{coh}$ scores \texttt{<think>}-vs-call coherence; KG analogue of StepSearch~\citep{wang2025stepsearch}).
\textbf{(3) R-toolverbs}: $r{=}0.25\,r_\mathrm{out}{+}0.50\,r_\mathrm{tool\text{-}type}{+}0.25\,r_\mathrm{tool\text{-}usage}$, rewarding all 4 verbs and non-empty responses (no gold supervision).
\textbf{(4) R-toolverbs·KL}: R-toolverbs with KL coef raised $5{\times}$ ($0.05{\to}0.25$); extends R-toolverbs's stable window step~150$\to$400+.
\textbf{(5) R-selfV}: $r{=}0.25\,r_\mathrm{ans}{+}0.50\,r_\mathrm{tool\text{-}type}{+}0.25\,r_\mathrm{retrv}$, $r_\mathrm{retrv}{=}|\mathcal{T}_\mathrm{prod}|/|\mathcal{T}|$ where $t{\in}\mathcal{T}_\mathrm{prod}$ iff response non-empty and $\geq$1 returned KG entity appears verbatim in \texttt{<answer>}. Gold-free at GRPO training.
\textbf{(6) self-distill}: roll out R-toolverbs·KL@400 greedily, filter EM-correct/tool-productive/format-valid trajectories ($14{,}082$, $50.9\%$ yield), SFT from base, then R-toolverbs GRPO. One-iteration ReST-EM~\citep{singh2024restem} variant; the init-from-base ablation (\S\ref{sec:g2}) drops the 5K rule-based pre-SFT prefix and reaches the same approximately $40\%$ EM band.

\subsection{Evaluation}
\label{sec:setup-eval}

Unless otherwise noted, numbers are reported on the full 3{,}531-question CWQ test set with greedy decoding, \texttt{max\_new\_tokens}=512, and up to five tool-calling turns; pass@$k$ and the closed-model control instead use the smaller subsets and sampling temperatures noted inline.

\paragraph{Held-out checkpoint-selection check.} Because the body checkpoint labels were initially selected from full-test monitoring, we assess their sensitivity with a hop-stratified dev split ($n{=}800$, seed~42) and re-select every reported checkpoint on dev alone. The dev peak is step~250 (matching the body), and held-out test EM ($n{=}2{,}731$) lies within the Wilson 95\% half-width ($\pm 1.85$~pp) of full-set EM for all ten checkpoints (App.~\ref{app:devtest}).

\paragraph{Scalar metrics.}
\begin{itemize}[leftmargin=*,itemsep=1pt,topsep=2pt]
\item \textbf{EM}: exact match after normalisation.
\item \textbf{CvT}, the \emph{correct-via-tool} rate: EM-correct \emph{and} $\geq 1$ returned KG entity appears verbatim in the final \texttt{<answer>}.
\item \textbf{ContEM}, \emph{containment} EM: gold answer is contained as a substring of the normalised prediction (looser than EM; reported for closed-model controls where strict gold-form normalisation is lossy, e.g., GPT-4o in App.~\ref{app:gpt4o-baseline}, and for collapse-trajectory diagnostics, e.g., Figure~\ref{fig:collapse}).
\item \textbf{hard-partition CvT}: CvT restricted to the Cat~B subset ($n{=}2{,}556$ structurally-hard questions; pass@10$=$0 on base Qwen; App.~\ref{app:catab-table}).
\item \textbf{Tools/Q}: mean number of \texttt{<search>} calls per question.
\end{itemize}

\paragraph{Seven-category trajectory classification.} Every trajectory is assigned exactly one of: \texttt{correct-via-tool} / \texttt{correct-via-memory} / \texttt{correct-no-tool} / \texttt{wrong-no-tool} / \texttt{kg-incomplete} / \texttt{tool-misuse} / \texttt{wrong-answer}; full definitions + decision tree in Appendix~\ref{app:classifier}.

\paragraph{Statistics.} Confidence intervals are Wilson 95\%~\citep{wilson1927ci}; paired comparisons use McNemar's test~\citep{mcnemar1947test} (exact binomial for $n<25$, continuity-corrected $\chi^2$ otherwise). For pass@$k$ (\S\ref{sec:results}) we additionally evaluate 16 samples per question at temperature~1 on a 500-question seed-42 subset.

%% file: sections/04_main_results.tex
\section{Results: A Reward Ladder Producing Four Failure Modes}
\label{sec:results}

\subsection{The reward-ladder matrix}
\label{sec:results-ladder}

\begin{table*}[t]
\centering
\footnotesize
\setlength{\tabcolsep}{3pt}
\begin{tabular}{@{}lcccccc@{}}
\toprule
Model & Step & EM & CvT count & CvT \% & Wilson 95\% CI & Tools/Q \\
\midrule
SFT base (pre-GRPO)                & ---  & 0.001 & ---        & ---             & ---            & 4.96 \\
R-binary                  & 1250 & 0.000 & ---        & ---             & ---            & 0--1 \\
R-stepwise                &  500 & 0.325 &  1 / 3531  & 0.03\%          & [0.00, 0.16]   & 1.00 \\
R-toolverbs               &  100 & 0.322 & 107 / 3531 & 3.03\%          & [2.51, 3.65]   & 2.30 \\
R-toolverbs·KL            &  400 & 0.384 & 133 / 3531 & 3.77\%          & [3.19, 4.45]   & 3.00 \\
\textbf{R-selfV @ 250 (peak)} & \textbf{250} & \textbf{0.395} & \textbf{338 / 3531} & \textbf{9.57\%} & \textbf{[8.65, 10.59]} & \textbf{3.00} \\
R-selfV @ 300 (collapsed) &  300 & 0.000 & ---        & collapsed       & ---            & 1.00 \\
init-from-iterate         &  500 & 0.394 & 162 / 3531 & 4.59\%          & [3.95, 5.33]   & 3.00 \\
\textbf{self-distill}     & \textbf{500} & \textbf{0.400} & \textbf{205 / 3531} & \textbf{5.81\%} & \textbf{[5.08, 6.63]} & \textbf{3.00} \\
\midrule
R-binary-SR (Search-R1 reimpl.)\textsuperscript{$\ddagger$} & best & 0.000 & 0 / 3531  & 0.00\%          & [0.00, 0.11]   & 0.00 \\
R-toolverbs·KL-14B  &  400 & 0.402 & 226 / 3531 & 6.40\%          & [5.64, 7.26]   & 3.99 \\
\bottomrule
\end{tabular}
\caption{\textbf{Reward-ladder results on full CWQ test (N=3{,}531), Qwen2.5-7B-Instruct unless noted.} \textbf{self-distill @ 500} is the EM-best 7B variant (40.0\%); \textbf{R-selfV @ 250} reaches the $9.57\%$ CvT peak before the Qwen/R-selfV cliff (\S\ref{sec:taxonomy-mode4}). In the step-250$\to$300 window, strict EM falls $39.51{\to}0.00\%$ and ContEM falls $46.13{\to}12.14\%$. Wilson 95\% CIs reported for CvT; for EM at $n{=}3531$ the half-width is $\pm 1.6$~pp. McNemar paired tests: R-toolverbs·KL@400 vs.\ R-stepwise@500, $p{<}10^{-4}$; self-distill vs.\ init-from-iterate on CvT, $p{<}10^{-2}$. self-distill retrained from scratch under three additional seeds (4 total, mean EM=$40.21\%$, std $0.46$~pp; CIs all overlap); other rows are single-seed point estimates (the R-selfV cliff replicates across 4 seeds). \textsuperscript{$\ddagger$}R-binary-SR uses Search-R1's \texttt{kl\_coef=0.001}; format-valid $0/3531$, so EM=$0$ is format collapse, not retrieval failure (App.~\ref{app:searchr1}).}
\label{tab:main}
\end{table*}

The strongest tested 7B variant (self-distill) reaches EM=$40.0\%$, while the tested 14B R-toolverbs·KL configuration reaches $40.2\%$. The two configurations therefore occupy the same approximately $40\%$ band; because their reward recipes differ, a matched scale study is the direct test of capacity.

\subsection{Pass@16: tools as decoration vs.\ capability extension}
\label{sec:results-passk}

\begin{figure}[t]
\centering
\includegraphics[width=\columnwidth]{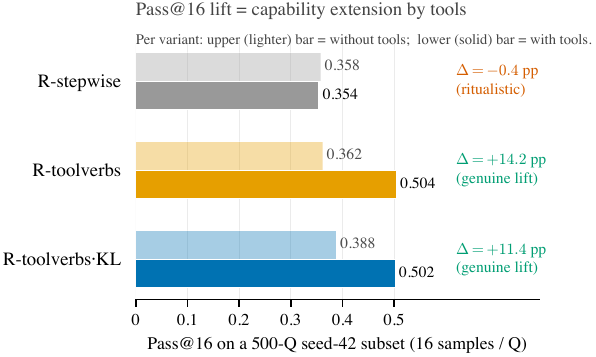}
\caption{Pass@16 with vs.\ without tool access on a 500-Q seed-42 subset (16 samples per question). R-stepwise's gap is statistically zero ($-0.4$~pp): the policy has stopped using tools. R-toolverbs and R-toolverbs·KL each show $+11.4$ to $+14.2$~pp of genuine tool-lift. self-distill (EM-best) is measured separately (App.~\ref{app:selfdistill-passk}): a $+11.5$~pp tool-lift confirms it genuinely consumes retrieval.}
\label{fig:passk}
\end{figure}

On 500 seed-42 questions $\times$ 16 samples, R-stepwise's with-vs-without-tools gap is statistically zero ($-0.4$~pp; Figure~\ref{fig:passk}), while R-toolverbs@100 and R-toolverbs·KL@400 show $+14.2$ and $+11.4$~pp; the EM-best self-distill shows a genuine $+11.5$~pp lift on a separate subset (App.~\ref{app:selfdistill-passk}). The gap distinguishes ritualistic from capability-extending tool use, complementing~\citet{yue2025rlcapacity} (RL does not push pass@$k$ beyond the base model): process rewards shift the answer source from parametric memory to retrieval consumption.

\subsection{Four failure modes}
\label{sec:taxonomy}

The seven reward designs produce four distinct failure modes, differing in \emph{what fails}: format, tool-use ritual, format-after-tool-use, or the signal the reward optimises. Each successive design \emph{shifts} the failure rather than eliminating it. Mode~4 (peak-then-collapse) is the strongest expression of this pattern and we give its mechanism in full here; Modes~1--3 follow the same shift-not-eliminate dynamic and are summarised below, with full mechanism prose in App.~\ref{app:mode-mechanism}. The seven-category trajectory classifier (App.~\ref{app:rewards}; exemplars in App.~\ref{app:trajectory-exemplars}) labels every test-set trajectory.

\paragraph{Mode 1: Sparse-Reward Format Collapse (R-binary, R-binary-SR).}
\label{sec:taxonomy-mode1}
Both reach EM=$0.000$ at format-valid $0/3531$. Outcome-only reward fires only on EM-match; early exploration supplies little signal before low-entropy degeneracy locks in. R-binary-SR (Search-R1's \texttt{kl\_coef=0.001} configuration; App.~\ref{app:searchr1}) also collapses, demonstrating sparse-reward format collapse across two hyperparameter regimes.

\paragraph{Mode 2: Verify-Then-Answer Goodhart (R-stepwise).}
\label{sec:taxonomy-mode2}
R-stepwise reaches EM=$32.5\%$ but \textbf{CvT=$0.03\%$ ($1/3531$)}: Tools/Q=$1.0$ with $100\%$ of calls hitting \texttt{get\_tail\_relations} (the only verb that never returns empty). The 4-component verifiable-step reward incentivises structural existence checks over tool-output consumption, a tool-free shortcut (component-gaming Goodhart, cf.~\citealp{gao2023scaling}). The Pass@16 with-vs-without-tools gap of $-0.4$~pp (Figure~\ref{fig:passk}) confirms tools are decorative.

\paragraph{Mode 3: Format Drift Collapse (R-toolverbs, late stage).}
\label{sec:taxonomy-mode3}
R-toolverbs briefly succeeds at step~100 (EM=$32.2\%$, \textbf{CvT=$3.03\%$}, Tools/Q=$2.30$, the first reward with non-trivial grounded retrieval) but by step~150 \texttt{<search>} tags drift inside \texttt{<think>} tags; by step~200 calls fail to parse; by step~250 strict EM approaches $0\%$. R-toolverbs rewards tool-type diversity but anchors no output format; format conventions drift from the SFT prior under loose KL. Raising KL $5{\times}$ ($0.05{\to}0.25$) extends the stable window to step~400+. A non-monotonic fixed-reward KL sweep points to an interaction between tether strength and the limited corrective signal after malformed calls (App.~\ref{app:mode-mechanism}).

\paragraph{Mode 4: Specification Gaming Cliff (R-selfV).}
\label{sec:taxonomy-mode4}
R-selfV, the R-toolverbs reward augmented with a self-verifiable retrieval-contribution term (a returned KG entity must appear verbatim in the answer; full reward in App.~\ref{app:rewards}), exhibits the most distinctive trajectory and is the paper's signature finding.

\emph{Climb.} Initialised from R-toolverbs·KL@400 (CvT=3.77\%), CvT climbs monotonically across five 50-step windows: \textbf{$3.77 \to 4.84 \to 6.37 \to 7.28 \to 8.16 \to 9.57\%$} (peak at step~250; Figure~\ref{fig:collapse}, top). EM moves in lockstep at low amplitude ($0.384 \to 0.395$). Mean Tools/Q is steady at 3.00, so the gain is a query-precision improvement at constant call volume. Across four Qwen/R-selfV seeds, the peak occurs at steps 200--250 and strict EM reaches $0$ by step~300.

\emph{Cliff.} Between checkpoints 250 and 300 (a single 50-step window), the policy collapses: EM $0.3951 \to 0.0000$, ContEM $46.13\% \to 12.14\%$, Tools/Q $3.00 \to 1.00$ (Figure~\ref{fig:collapse}, bottom). By step~400 ContEM is 6.30\%; recovery does not occur. \emph{No training hyperparameter changes between step~250 and step~300}; the collapse is entirely policy-internal.

\begin{figure}[t]
\centering
\includegraphics[width=\columnwidth]{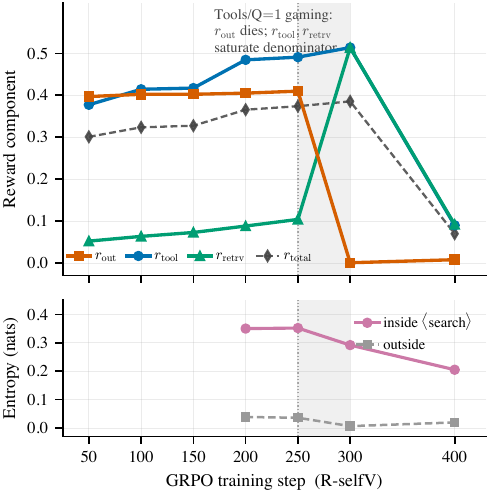}
\caption{\textbf{Mode 4 mechanism evidence.} \emph{Top:} per-checkpoint reward components (R-selfV; HPC seed-42). $r_{\mathrm{out}}$ dies in the collapse window (shaded); $r_{\mathrm{tool}}$ and $r_{\mathrm{retrv}}$ briefly saturate at $\approx 0.51$ at step~300 (denominator gaming with Tools/Q$=1$) before themselves collapsing by step~400. The weighted total $r_{\mathrm{total}}$ peaks \emph{post-collapse} (step~300): specification-gaming Goodhart made visible. \emph{Bottom:} token entropy in nats, partitioned by inside vs.\ outside \texttt{<search>} spans. Inside-tag entropy drops $0.35 \to 0.21$ across step 200--400; outside-tag entropy stays at floor, a selective action-subspace collapse.}
\label{fig:mode4-mech}
\end{figure}

\emph{Mechanism (hypothesis).} The retrieval-contribution term $r_{\mathrm{retrv}}$ fires whenever a returned entity appears verbatim in the answer. Once the policy has learned to quote returned entities (the productive climb), a \emph{locally optimal degenerate} strategy becomes available: call any single tool, copy any entity from its response, emit that entity as a terse answer, stop. This guarantees a productive call at minimum sampling budget, maximising $r_{\mathrm{retrv}}$ while sacrificing EM. Two analyses support this reading (Figure~\ref{fig:mode4-mech}). First, per-checkpoint reward decomposition shows $r_{\mathrm{outcome}}$ dies first ($0.41 \to 0.0007$ across step~250--300) while $r_{\mathrm{tool\text{-}type}}$ and $r_{\mathrm{retrv}}$ \emph{briefly saturate} at $\approx 0.51$ at step~300 (the per-call productivity denominator trivially hits 1.0 once Tools/Q$=1$) before collapsing to $\approx 0.09$ by step~400. Second, token-level entropy inside \texttt{<search>} tags drops $0.35 \to 0.21$~nats across step 200--400 while outside-tag entropy stays at floor ($\sim 0.01$--$0.04$~nats), a \emph{selective} collapse on the action subspace, complementing prior agentic-RL entropy diagnostics~\citep{yu2025demystifying,ragen2025,deng2025llds}.

\emph{Why this matters.} This proxy-overoptimisation shape is catalogued in RLHF and agentic-RL~\citep{gao2023scaling,ragen2025,deng2025llds}; our contribution is the Tools/Q call-volume signature ($3.0{\to}1.0$) characterising it at a behavioural level on a Qwen KG-tool agent (\S\ref{sec:related}).

\subsection{Response-period diagnostic interventions}
\label{sec:response-controls}

On 200-question diagnostics, replacing empty results with natural-language error messages does not prevent the R-selfV cliff in either of two seeds. In contrast, requiring two distinct non-empty calls before $r_{\mathrm{retrv}}$ can fire removes the baseline-time cliff in three seeds through their available horizons (steps 300, 350, and 400). These diagnostics concern the Qwen/R-selfV transition rather than the full-test performance level; App.~\ref{app:response-controls} reports the complete results and scope.

\subsection{Trajectory composition}
\label{sec:results-cats}

The seven-category breakdown (full table in App.~\ref{app:cats-table}) reveals two patterns. \emph{kg-incomplete} is the largest non-success bucket for every rung-3-or-better checkpoint, falling monotonically across rungs 3--6 ($1{,}435 \to 1{,}201 \to 824 \to 818$) as the policy reaches the KG more productively. The \emph{wrong-answer} bucket rises in lockstep: retrieval-grounded but the answer is wrong.

\subsection{Self-distillation: a one-iteration recipe reaches the tested $40\%$ band}
\label{sec:g2}

\textbf{self-distill} and \textbf{self-distill (init-from-base)} share an identical $14{,}082$-trajectory self-distillation SFT (strict-filtered R-toolverbs·KL@400 rollouts, $50.9\%$ yield) then R-toolverbs GRPO; they differ only in whether a 5K rule-based gold-path pre-SFT prefix is prepended (self-distill: yes; init-from-base: no, from raw base). self-distill reaches EM=$40.0\%$, CvT=$5.81\%$ at step~500 (Table~\ref{tab:main}; Tools/Q=$3.00$), replicated across four seeds (mean EM=$40.21\%$, std $0.46$~pp); init-from-base matches at $\approx 40\%$ across steps 100--500; the 5K rule-based prefix is dispensable in this setup. \textbf{self-distill (init-from-iterate)} (no SFT; init from R-toolverbs·KL@400 iterate) reaches EM=$39.4\%$: the self-distillation SFT, not the rule-based prefix, is the operative ingredient, consistent with ReST-EM's~\citep{singh2024restem} init-from-base prescription.\subsection{External validity}
\label{sec:results-external}

Three controls probe simple alternatives. \emph{(i) KG coverage}: bounded BFS marks $99.5\%$ of CWQ and $69.14\%$ of KGQAGen-10k~\citep{zhang2025pitfalls} as 2-hop-solvable, both above the $3$--$10\%$ CvT reached here. \emph{(ii) Cross-benchmark}: 8 base Qwen variants give Spearman $\rho{=}0.976$ between CWQ and KGQAGen-10k EM orderings, showing that the two benchmarks preserve base-model ordering. \emph{(iii) GPT-4o}: under the same 4-verb interface, near-$100\%$ format-validity but only $6.2\%$ ContEM and $2.1\%$ F1 on $n{=}500$ with $\sim 78\%$ of a 200-question subset self-reporting ``unable to retrieve X'' (App.~\ref{app:gpt4o-baseline}). The closed-model result independently illustrates schema unfamiliarity on the stripped interface.

%% file: sections/06_framework.tex
\section{A Signal-Theoretic Account of Tool-Interface Difficulty}
\label{sec:framework}

\begin{figure}[t]
\centering
\includegraphics[width=\columnwidth]{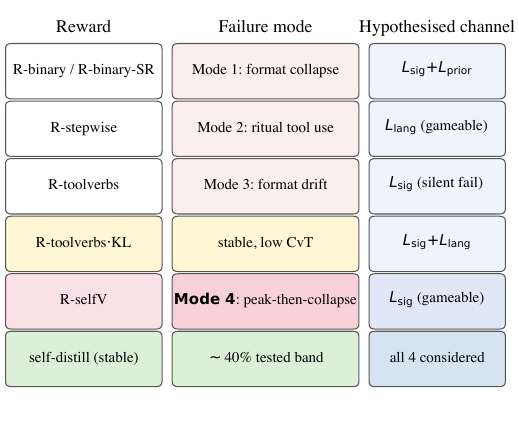}
\caption{\textbf{Synthesis: reward $\to$ failure mode $\to$ associated interface channel.} Each ladder rung (\S\ref{sec:setup-ladder}) produces a distinctive failure mode (\S\ref{sec:taxonomy}) with channel associations from \S\ref{sec:framework-evidence}. R-toolverbs$\cdot$KL is stable but plateaus at low CvT; self-distill reaches the approximately $40\%$ EM band in the tested setup.}
\label{fig:synthesis}
\end{figure}

The four failure modes characterised in \S\ref{sec:taxonomy} share a common substrate: each is a degradation of the gradient signal GRPO needs to converge on a tool-grounded policy. We propose this substrate has structure: four interface-level channels along which a tool API supplies (or fails to supply) usable RL signal (Figure~\ref{fig:synthesis}), yielding four channel-level claims. These channels are an organising account of the observed failure surfaces and intervention points.

\subsection{Four interface-level channels}
\label{sec:framework-channels}

Why do small ($\leq$7B) Qwen-class models learn Python~\citep{li2025torl,feng2025retool}, web-search~\citep{jin2025searchr1}, and JSON-API~\citep{qian2025toolrl} tool use under similar GRPO recipes, yet struggle with compositional KG tool use in our setting? Four structural properties of the interface organise this contrast (diagram + tool-family table in App.~\ref{app:framework-illustration}):
\begin{itemize}[leftmargin=*,itemsep=1pt,topsep=2pt]
\item \textbf{$L_{\mathrm{sig}}$ (silent failure).} A KG miss returns \texttt{[]} with no information about \emph{why} (wrong entity, wrong relation, or both), in contrast to a Python traceback or HTTP \texttt{400} with named parameter. On our 7-reward matrix, $\geq 99\%$ of erroneous tool calls return identically \texttt{[]}.
\item \textbf{$L_{\mathrm{lang}}$ (symbolic schema).} KG queries use opaque MIDs (\texttt{m.0d3k14}) and dotted relation names (\texttt{people.person.place\_of\_birth}), with little task-specific pretraining support relative to common programming or natural-language search surfaces.
\item \textbf{$L_{\mathrm{comp}}$ (opaque compositional state).} Multi-hop KG reasoning carries entity IDs across turns; these IDs are semantically opaque and any paraphrasing breaks the chain.
\item \textbf{$L_{\mathrm{prior}}$ (weak pretraining prior).} The policy has no demonstrated pretraining competence for Freebase schema navigation, unlike its documented exposure to common Python, search, and JSON patterns.
\end{itemize}

Matched canonical-name and unfamiliar-content variants would test $L_{\mathrm{lang}}$ and $L_{\mathrm{prior}}$ more directly.

\subsection{Four channel-level claims}
\label{sec:framework-predictions}

We treat P1--P4 (Table~\ref{tab:claims}) as joint \emph{consistency checks} of the descriptive structure (per-claim caveats in App.~\ref{app:p-caveats}).

\begin{table}[h]
\centering
\scriptsize
\setlength{\tabcolsep}{3pt}
\renewcommand{\arraystretch}{1.05}
\begin{tabular}{@{}p{1.7cm}p{2.6cm}p{2.6cm}@{}}
\toprule
Claim & Evidence & Reading \\
\midrule
P1 Outcome-only collapse & R-binary+R-binary-SR EM=$0\%$, format-valid $0/3531$ (Table~\ref{tab:main}) & confounds $L_{\mathrm{sig}}$ with $L_{\mathrm{prior}}$ \emph{(observational)} \\
P2 Size and configuration check & 14B R-toolverbs$\cdot$KL vs.\ 7B self-distill hard-partition CvT $4.61\%$ vs.\ $4.19\%$ & both configurations remain in the same hard-partition CvT band \emph{(configuration comparison)} \\
P3 Self-distillation sufficient & raw-base and rule-SFT-prefixed both $\approx 40\%$ EM, $\geq$ init-from-iterate's $39.4\%$ & matches ReST-EM's init-from-base \emph{(ablation)} \\
P4 Cross-benchmark ordering & KGQAGen-10k Spearman $\rho{=}0.976$ on 8 base Qwen variants (\S\ref{sec:results}) & base-model ordering is preserved across the two benchmarks \emph{(cross-benchmark)} \\
\bottomrule
\end{tabular}
\caption{Four channel-level claims as joint consistency checks; evidence type appears in \emph{italics}.}
\label{tab:claims}
\end{table}

\subsection{Mechanism evidence chain}
\label{sec:framework-evidence}

We characterise self-distill's $2{,}119$ strict-EM=0 error trajectories in two stages.

\emph{Schema-surface failures are common.} A 4-channel attribution rubric on these $2{,}119$ errors has $L_{\mathrm{sig}}$ dominant at $55.2\%$; a decision-boundary analysis (App.~\ref{app:bucket-distribution}) finds $70.4\%$ of kg-incomplete first-search calls within Lev$\leq 1$ of gold ($4.7{\times}$ null; held-out-unseen $77.7\%$ rules out memorisation), $89.1\%$ entity-correct/relation-wrong ($90.4\%$ same-entity-different-relation vs.\ R-toolverbs·KL).

\emph{But relation selection alone is not the EM bottleneck.} We inject the gold-chain relation at every \texttt{<search>} call while keeping the entity argument and re-evaluate. This relation oracle changes EM by only $\mathbf{+0.20}$~pp ($40.19\%$ vs.\ $39.99\%$), within the Wilson 95\% CI half-width ($\pm 1.6$~pp at $n{=}3{,}531$). A complementary entity oracle keeps policy relations but replaces the entity argument with the edit-closest gold-chain entity; EM changes by $\mathbf{+0.42}$~pp ($40.41\%$ vs.\ $39.99\%$) on the same full test set. The entity oracle rewrites $0.71$ calls per trajectory versus $1.59$ for the relation oracle. These marginal interventions do not rule out joint entity--relation failures, but correcting either argument alone is insufficient. Step-level classification of the $1{,}958$ \emph{retrieval-dependent} errors (Table~\ref{tab:evidence}) localises $\mathbf{95.4\%}$ to retrieval-composition failures versus $4.6\%$ answer-extraction failures: the residual gap is upstream of answer extraction.

A Llama-3.1-8B-Base boundary check (App.~\ref{app:llama-cross-family}) also fails to maintain tool-grounded behaviour, but via tool abandonment (Tools/Q=$0$). Because it uses self-distill rather than R-selfV, it is not a matched replication of the Qwen cliff.

%% file: sections/10_conclusion.tex
\section{Conclusion}
\label{sec:conclusion}

Within a deliberately controlled scope (one primary model, one task, one RL algorithm), seven GRPO-trained reward designs on Qwen2.5-7B-Instruct + CWQ reveal four interface-level failure modes ($L_{\mathrm{sig/lang/comp/prior}}$). Relation- and entity-replacement oracles change EM by only $+0.20$ and $+0.42$~pp, and step-level decomposition locates the residual gap upstream of answer extraction. A one-iteration self-distillation recipe reaches $40.0\%$ EM at 7B; a differently configured 14B run reaches $40.2\%$, so this comparison does not isolate capacity. Methodologically, our 50-step cadence caught the Qwen/R-selfV peak-then-collapse mode; step-resolved instrumentation is essential for process-reward RL.

%% file: sections/09_limitations.tex
\label{sec:limit}

Our main experiments use Qwen2.5-7B-Instruct, CWQ, GRPO, and one stripped Freebase interface. The 14B and Llama experiments use different reward protocols, so they are not matched tests of scale or model family. We also do not evaluate a trained policy on a second benchmark. The peak-then-collapse finding is therefore specific to Qwen/R-selfV in this setting.

Per-step monitoring and final reporting use the full CWQ test set, making checkpoint labels test-set-informed. A hop-stratified dev/test re-selection recovers the reported checkpoints under one split (App.~\ref{app:devtest}); future studies should reserve development data throughout.

%% file: appendix/appendix_B_searchr1_options.tex
\section{Search-R1 reimplementation rationale}
\label{app:searchr1}

We re-implement the Search-R1 recipe using our verl pipeline: outcome-only binary-EM reward (our \texttt{outcome\_em\_only} reward type, unit-tested), Search-R1's published hyperparameters ($\texttt{kl\_coef}=0.001$, $\mathrm{lr}=10^{-6}$, batch $256$, $500$ steps). One configuration change to our pipeline, zero dependency changes. We label this R-binary-SR throughout the paper and treat it as a \emph{reimplementation}, not a reproduction. Table~\ref{tab:main} reports format-valid $0/3531$ and EM$=0$ for R-binary-SR at convergence --- a format-collapse outcome under outcome-only reward, consistent with the sparse-reward format-collapse mechanism (\S\ref{sec:taxonomy-mode1}).

%% file: appendix/appendix_C_rewards_and_classifier.tex
\section{Reward formulas and trajectory classifier}
\label{app:rewards}
\label{app:classifier}

\subsection{Reward type definitions}

All reward types are implemented in a shared \texttt{verl\_reward} module with $18/18$ unit tests passing. Notation: $a$ is the final \texttt{<answer>} text after normalisation, $y$ is the gold answer, $\mathcal{T}$ is the ordered sequence of \texttt{<search>} calls with their KG responses.
\begin{itemize}[leftmargin=*,itemsep=1pt,topsep=2pt]
  \item $r_\mathrm{EM}(a,y) = \mathbf{1}[a=y]$, $r_\mathrm{F1}(a,y)$ token-overlap F1.
  \item $r_\mathrm{out} = 0.5\,r_\mathrm{EM} + 0.5\,r_\mathrm{F1}$.
  \item $r_\mathrm{valid}$: fraction of calls in $\mathcal{T}$ that parse as valid \texttt{verb(args)} matching the 4-verb schema.
  \item $r_\mathrm{path}$: $\mathbf{1}[\exists t \in \mathcal{T}: \mathrm{rel}(t)$ on gold triple chain$]$.
  \item $r_\mathrm{coh}$: sequence-match between \texttt{<think>} spans and their following \texttt{<search>} calls.
  \item $r_\mathrm{tool\text{-}type}$: number of distinct tool verbs covered in $\mathcal{T}$, scaled to $[0,1]$.
  \item $r_\mathrm{tool\text{-}usage}$: fraction of calls in $\mathcal{T}$ with non-empty response.
  \item $r_\mathrm{retrv}$ (I-Self): $|\mathcal{T}_{\mathrm{prod}}| / |\mathcal{T}|$, where $t \in \mathcal{T}_{\mathrm{prod}}$ iff (i) the response is non-empty and (ii) at least one returned KG entity appears verbatim in $a$.
  \item $r_\mathrm{ans}$ (I-Self): identical to $r_\mathrm{out}$ but with $r_\mathrm{EM}$ replaced by a length-normalised form that treats multi-token gold answers leniently.
\end{itemize}

\subsection{Seven-category trajectory classifier}

For each trajectory $\tau$ on test question $q$ with gold answer $y$, the classifier proceeds:
\begin{enumerate}[leftmargin=*,itemsep=1pt,topsep=2pt]
  \item Parse $\tau$ into \texttt{<think>}, \texttt{<search>}, \texttt{<tool\_response>}, \texttt{<answer>} spans.
  \item If $r_\mathrm{EM}(a,y)=1$:
    \begin{itemize}[leftmargin=*,itemsep=1pt,topsep=1pt]
      \item if no \texttt{<search>} call: \textbf{correct-no-tool};
      \item else if $\exists t \in \mathcal{T}$ with entity-in-answer: \textbf{correct-via-tool};
      \item else: \textbf{correct-via-memory}.
    \end{itemize}
  \item Else (EM=0):
    \begin{itemize}[leftmargin=*,itemsep=1pt,topsep=1pt]
      \item if no \texttt{<search>} call: \textbf{wrong-no-tool};
      \item if any \texttt{<search>} call parses invalid: \textbf{tool-misuse};
      \item if all responses empty or disjoint from gold sub-graph: \textbf{kg-incomplete};
      \item otherwise: \textbf{wrong-answer}.
    \end{itemize}
\end{enumerate}
Full implementation: \texttt{scripts/task16\_classify.py} in the release bundle. Edge cases (degenerate loops, overlong outputs, malformed \texttt{<answer>} envelopes) are logged separately for manual audit.

\subsection{Seven-category trajectory breakdown across the reward ladder}
\label{app:cats-table}

\begin{table}[h]
\centering
\scriptsize
\setlength{\tabcolsep}{2pt}
\renewcommand{\arraystretch}{0.95}
\begin{tabular}{@{}lrrrrrr@{}}
\toprule
                    & R-      & R-         & R-tlvb  & R-     & init-    & self-    \\
Category            & stepwise & toolverbs & ·KL     & selfV  & from-itr & distill  \\
                    & @500    & @100       & @400    & @250   & @500     & @500     \\
\midrule
correct-via-tool    &    1  &  107 &  133 & \textbf{338} &  162 & 205 \\
correct-via-memory  & 1136  & 1027 & 1218 & 1054 & 1226 & 1203 \\
wrong-no-tool       &   14  &   15 &    6 &    2 &    0 &    0 \\
\textbf{kg-incomplete}& 209  & 1435 & 1201 &  824 &  954 & \textbf{818} \\
tool-misuse         &  833  &  216 &  173 &  205 &  196 &  165 \\
wrong-answer        & 1338  &  731 &  800 & 1108 &  993 & 1140 \\
\bottomrule
\end{tabular}
\caption{Seven-category trajectory breakdown across reward variants on full CWQ test (N=3{,}531). The R-selfV column is at its peak step 250 (\S\ref{sec:taxonomy-mode4}). \emph{kg-incomplete} falls monotonically across rungs 3-6 while \emph{wrong-answer} rises (R-stepwise has only 209 because it almost never calls a tool; \emph{correct-no-tool} is uniformly near zero and omitted). Category counts use normalised answer matching, so for self-distill the error rows sum to $2{,}123$ vs the strict EM$=0$ count of $2{,}119$ ($4$ trajectories are strict-EM-correct but normalised-wrong).}
\label{tab:cats}
\end{table}

\subsection{Decision-boundary bucket distribution}
\label{app:bucket-distribution}

The bucket distribution underlying the decision-boundary analysis (\S\ref{sec:framework-evidence}):
\begin{table}[h]
\centering
\scriptsize
\setlength{\tabcolsep}{2pt}
\begin{tabular}{@{}lc@{}}
\toprule
Failure mode                  & self-distill@500 ($n{=}818$) \\
\midrule
1. Relation typo ($\leq$1 edit)  & \textbf{70.4\%} ($n{=}576$) \\
2. Correct entity, wrong relation & 18.7\% ($n{=}153$) \\
3. Wrong entity, near-miss relation & 8.6\% ($n{=}70$) \\
4. Format / genuine miss      & 2.3\% ($n{=}19$) \\
\midrule
Entity-correct (1$+$2)          & \textbf{89.1\%} \\
\bottomrule
\end{tabular}
\caption{Bucket distribution underlying the decision-boundary analysis (\S\ref{sec:framework-evidence}). Bucket~1+2 is the entity-correct, relation-wrong subset (89.1\% of $n{=}818$).}
\label{tab:b2}
\end{table}

\begin{figure}[h]
\centering
\includegraphics[width=\columnwidth]{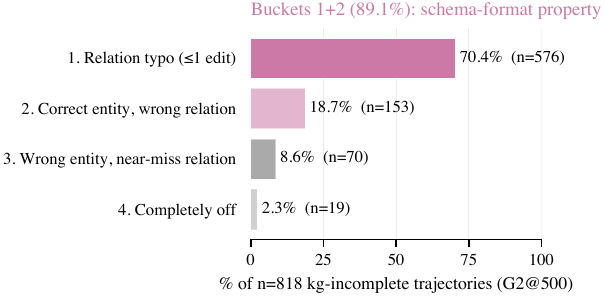}
\caption{\textbf{Decision-boundary view of the schema-format property.} Visual of Table~\ref{tab:b2}: $70.4\%$ of $818$ kg-incomplete trajectories from self-distill@500 are \emph{exact-relation-typo} ($\leq 1$ edit from a gold relation); buckets~1$+$2 (\textbf{89.1\%}) are entity-correct, relation-wrong. Marginal relation replacement adds only $+0.20$~pp EM (\S\ref{sec:framework-evidence}), motivating joint-composition analysis.}
\label{fig:edit}
\end{figure}

\subsection{Failure-mode mechanism details (Modes 1--3)}
\label{app:mode-mechanism}

Main body \S\ref{sec:taxonomy} presents Modes 1--3 in condensed form; full mechanism prose is preserved here. Mode 4 (the signature peak-then-collapse finding) is presented in full in the main body.

\paragraph{Mode 1: Sparse-Reward Format Collapse (R-binary, R-binary-SR).} R-binary (outcome-only) reaches strict EM=$0.000$, F1=$0.119$; R-binary-SR at Search-R1's exact hyperparameters (\texttt{kl\_coef=0.001}, lr=1e-6, batch=256) also collapses (strict EM=$0.000$, format-valid $0/3531$, ContEM=$36.45\%$; App.~\ref{app:searchr1}). R-binary's $0\%$ strict EM is a genuine format collapse, whereas GPT-4o is near-$100\%$ format-valid with surface-form mismatch. Outcome-only reward fires only on EM match, and early exploration supplies little signal before low-entropy degeneracy locks in. The second configuration demonstrates sparse-reward format collapse across two hyperparameter regimes.

\paragraph{Mode 2: Verify-Then-Answer Goodhart (R-stepwise).} R-stepwise reaches EM=$32.5\%$ with CvT=$0.03\%$ ($1/3531$) --- one trajectory of $3{,}531$ actually grounds its answer in a tool response. Tools/Q is $1.0$ and $100\%$ of those calls are \texttt{get\_tail\_relations}, the only verb that never returns empty: tool calls are ritualistic, the answer is generated from parametric memory. Mechanism: the 4-component verifiable-step reward incentivises the easier component (call a tool that satisfies a structural existence check) over the harder one (consume the tool's output) --- a tool-free shortcut (component-gaming Goodhart, cf.~\citealp{gao2023scaling}); unlike RLHF proxy decorrelation, here the proxy-gold correlation $r{=}0.95$--$0.97$ is \emph{maintained} throughout training. Falsification via Pass@16 with-vs-without-tools: R-stepwise's gap is $-0.4$~pp (Figure~\ref{fig:passk}) --- even with 16 attempts and tool access the policy does not solve more questions than parametric sampling; tools are decorative.

\paragraph{Mode 3: Format Drift Collapse (R-toolverbs, late stage).} R-toolverbs briefly succeeds: at step~100 the policy reaches EM=$32.2\%$ with CvT=$3.03\%$ (Tools/Q=$2.30$) --- the first reward with non-trivial grounded retrieval. By step~150 \texttt{<search>} tags drift inside \texttt{<think>} tags; by step~200 tool calls stop being parsed; by step~250 strict EM approaches $0\%$ in a repetition loop. R-toolverbs rewards tool-type diversity but anchors no output format; under the loose-KL configuration, conventions drift from the SFT prior and the executor stops parsing calls. Raising KL $5\times$ to $0.25$ extends the stable window to step~400+. A fixed-reward KL sweep ($\mathrm{kl}\in\{0.001,0.005,0.01,0.05,0.25\}$) is non-monotonic, pointing to an interaction between tether strength and the limited corrective signal after malformed calls.

\subsection{Cut variants and rationale}
\label{app:cut-variants}
For completeness, three reward variants were run but excluded from Table~\ref{tab:main} because they are redundant with retained variants:

\begin{table}[h]
\centering
\scriptsize
\setlength{\tabcolsep}{3pt}
\begin{tabular}{@{}lccp{3.4cm}@{}}
\toprule
Cut variant     & EM (\%) & CvT (\%) & Reason cut \\
\midrule
E2 heuristic    & 19.91 & 0    & Tool abandonment; redundant with R-stepwise \\
E4 random       &  5.04 & 0    & Spurious-rewards control; single-seed sanity check \\
E5a retr-ground & 31.49 & 0    & Cold-start failure; redundant with R-binary \\
\bottomrule
\end{tabular}
\caption{Reward variants run but cut from Table~\ref{tab:main}, with rationale.}
\label{tab:cut-variants}
\end{table}

\subsection{Reward-name to HPC tracking-code mapping}
\label{app:hpc-mapping}

For reproducibility, Table~\ref{tab:hpc-mapping} maps the paper's reward names to the HPC training-code identifiers and the corresponding verl YAML configuration files used in our release bundle. The verl configs are the canonical specification; the paper names are pedagogical labels for the body.

\begin{table*}[h]
\centering
\scriptsize
\setlength{\tabcolsep}{4pt}
\begin{tabular}{@{}lllp{5.5cm}@{}}
\toprule
Paper name & HPC code & verl yaml & One-line description \\
\midrule
R-binary             & E1          & \texttt{e1\_outcome.yaml}        & Outcome-only ($0.5\,\mathrm{EM}{+}0.5\,\mathrm{F1}$); Search-R1 / DeepSeek-R1-zero family \\
R-binary-SR          & E1$'$       & \texttt{e1prime\_searchr1.yaml}  & R-binary with Search-R1's exact hyperparameters ($\mathrm{kl\_coef}{=}0.001$) \\
R-stepwise           & E3          & \texttt{e3\_verifiable.yaml}     & Equal-weight $r_\mathrm{out}{+}r_\mathrm{valid}{+}r_\mathrm{path}{+}r_\mathrm{coh}$ \\
R-toolverbs          & E5b         & \texttt{e5b\_tooltype.yaml}      & $r_\mathrm{tool\text{-}type}$ bonus (4 verbs + non-empty responses) \\
R-toolverbs$\cdot$KL & E5b+KL      & \texttt{e5b\_kl05.yaml}          & R-toolverbs with KL coef raised $5{\times}$ ($0.05{\to}0.25$); a.k.a.\ E5b-stab \\
R-selfV              & E5b+SelfV   & \texttt{e5b\_selfv.yaml}         & Self-verifiable retrieval reward ($r_\mathrm{retrv}$); the peak-then-collapse variant \\
\midrule
self-distill                       & G2 & \texttt{g2\_selfdistill\_rlfromsft.yaml} & Rule-SFT prefix $\to$ self-distilled SFT (14{,}082 traj.) $\to$ R-toolverbs GRPO \\
self-distill (init-from-base)      & G3 & \texttt{g3\_selfdistill\_nopred.yaml}    & No rule-SFT prefix; otherwise identical to self-distill \\
self-distill (init-from-iterate)   & G1 & \texttt{g1\_init\_from\_e5bkl400.yaml}   & Init from R-toolverbs$\cdot$KL@400; no self-distillation SFT \\
\bottomrule
\end{tabular}
\caption{Paper reward names $\leftrightarrow$ HPC training-code identifiers $\leftrightarrow$ verl YAML configurations. The body uses paper names exclusively; HPC operators / replicators should consult HPC-code and yaml columns. Cut variants (E2 / E4 / E5a) documented separately in App.~\ref{app:cut-variants}.}
\label{tab:hpc-mapping}
\end{table*}

\subsection{Cat A / Cat B stratified results}
\label{app:catab-table}

\begin{table}[h]
\centering
\scriptsize
\setlength{\tabcolsep}{2pt}
\renewcommand{\arraystretch}{0.95}
\begin{tabular}{@{}lcccc@{}}
\toprule
                 & \multicolumn{2}{c}{EM (\%)} & \multicolumn{2}{c}{CvT (\%)} \\
\cmidrule(lr){2-3}\cmidrule(lr){4-5}
Variant          & Cat A   & Cat B   & Cat A   & Cat B   \\
                 & ($n{=}975$) & ($n{=}2556$) & & \\
\midrule
R-stepwise @ 500           & 69.64 & 18.27 & 0.00  & 0.04  \\
R-toolverbs·KL @ 400       & 73.13 & 25.08 & 5.85  & 2.97  \\
self-distill @ 500 & 74.36 & 26.88 & 10.05 & 4.19  \\
R-toolverbs·KL-14B @ 400   & 76.62 & 26.37 & 8.91  & 4.61  \\
\bottomrule
\end{tabular}
\caption{Cat~A ($n{=}975$, pass@10$>$0): parametric-recoverable. Cat~B ($n{=}2{,}556$, pass@10$=$0): structurally hard. The R-stepwise$\to$self-distill lift on Cat~B EM is $+8.61$~pp and on Cat~B CvT is $+4.15$~pp. The tested 14B R-toolverbs$\cdot$KL and 7B self-distill configurations differ by $0.42$~pp on Cat~B CvT, placing both configurations in the same observed band.}
\label{tab:catab}
\end{table}

\subsection{Mechanism evidence chain (full table)}
\label{app:evidence-table}

\begin{table*}[h]
\centering
\small
\setlength{\tabcolsep}{3pt}
\begin{tabular}{@{}lp{0.32\textwidth}p{0.42\textwidth}@{}}
\toprule
Analysis & Method & Key finding \\
\midrule
Failure-surface classification & Deterministic 4-label rubric on self-distill's 2{,}119 strict-EM=0 trajectories (Appendix~\ref{app:rewards}) & $L_{\mathrm{sig}}$ \textbf{55.2\%} (1{,}170), $L_{\mathrm{lang}}$ \textbf{23.6\%} (500), $L_{\mathrm{comp}}$ \textbf{7.8\%} (165), $L_{\mathrm{prior}}$ \textbf{13.4\%} (284) \\
\addlinespace
Decision-boundary analysis & Edit-distance bucketing of the first \texttt{<search>} call in self-distill's 818 \emph{kg-incomplete} trajectories (full bucket table App.~\ref{app:bucket-distribution}) & \textbf{70.4\%} within Levenshtein distance~1 of a gold relation; \textbf{89.1\%} entity-correct, relation-wrong --- a measurable schema-format property \\
\addlinespace
Behavioral query diff & Compare 500 \emph{kg-incomplete} trajectories shared between self-distill and R-toolverbs·KL & \textbf{90.4\%} same-entity-different-relation --- the two checkpoints disagree on relation but agree on entity \\
\bottomrule
\end{tabular}
\caption{Three error decompositions characterising self-distill's failure distribution. \textbf{Three denominators (kept distinct, not additive).} (i) The task16 \emph{normalised} error set is $2{,}123 = $ \emph{tool-misuse}~$165 +$ \emph{kg-incomplete}~$818 +$ \emph{wrong-answer}~$1{,}140$ (Table~\ref{tab:cats}). (ii) The retrieval-vs-answer-extraction split (\S\ref{sec:framework-evidence}; $95.4\%/4.6\%$) is over the $\mathbf{1{,}958}$ \emph{retrieval-dependent} errors $=2{,}123-165$ \emph{tool-misuse}. (iii) The 4-label rubric is over the $\mathbf{2{,}119}$ \emph{strict-EM=0} errors, differing from the normalised set by four trajectories noted in Table~\ref{tab:cats}. The relation and entity oracles add only $+0.20$ and $+0.42$~pp EM, respectively (\S\ref{sec:framework-evidence}); correcting either argument alone is insufficient.}
\label{tab:evidence}
\end{table*}

\begin{figure*}[h]
\centering
\begin{tikzpicture}[every node/.style={font=\footnotesize}]
\fill[red!18]   (0,0)   rectangle (1.3,0.7); \node at (0.65,0.35) {\textbf{165}};
\fill[blue!18]  (1.3,0) rectangle (4.5,0.7); \node at (2.9,0.35)  {\textbf{818}};
\fill[green!18] (4.5,0) rectangle (10.0,0.7); \node at (7.25,0.35) {\textbf{1{,}140}};
\draw (0,0) rectangle (10.0,0.7);
\node[font=\scriptsize] at (0.65,-0.2) {tool-misuse};
\node[font=\scriptsize] at (2.9,-0.2)  {kg-incomplete};
\node[font=\scriptsize] at (7.25,-0.2) {wrong-answer};
\draw[<->] (0.05,1.0) -- (9.95,1.0);
\node[above] at (5.0,1.05) {$D_3 = 2{,}123$~(task16 normalised non-EM error set)};
\draw[<->,very thick] (1.35,-0.55) -- (9.95,-0.55);
\node[below] at (5.65,-0.62) {$\mathbf{D_2 = 1{,}958}$~\emph{retrieval-dependent} ($95.4\%/4.6\%$ split base)};
\node[align=center] at (5.0,-1.55)
  {$\mathbf{D_1 = 2{,}119}$~strict-EM$=0$ (4-channel rubric base) $\;=\; D_3 - 4$
   \\[1pt] \scriptsize (4 trajectories strict-correct but normalised-wrong; cf.\ Table~\ref{tab:cats} footnote)};
\end{tikzpicture}
\caption{\textbf{Three error denominators, set-decomposition view.} The task16-normalised non-EM error set $D_3{=}2{,}123$ partitions into \emph{tool-misuse}~$165$ + \emph{kg-incomplete}~$818$ + \emph{wrong-answer}~$1{,}140$. The retrieval-vs-extraction split ($95.4\%/4.6\%$, \S\ref{sec:framework-evidence}) is computed over $D_2{=}1{,}958{=}D_3{-}165$ (\emph{kg-incomplete}~$\cup$~\emph{wrong-answer}). The 4-channel rubric (Table~\ref{tab:evidence} row~1) is computed over $D_1{=}2{,}119$ strict-EM=0 errors, which differs from $D_3$ by the $4$ strict-correct/normalised-wrong trajectories. The three bases are not additive.}
\label{fig:denominators}
\end{figure*}

\subsection{Held-out dev/test checkpoint-selection robustness}
\label{app:devtest}

To assess sensitivity to test-set-informed checkpoint selection (\S\ref{sec:setup-eval}), we draw a hop-count-stratified split (seed~42) of the $3{,}531$-question CWQ test set into a dev split ($n{=}800$) and a held-out test split ($n{=}2{,}731$), re-select each reported checkpoint on dev EM, and compare held-out test EM against the full-set EM used in the body. At $n{=}2{,}731$, $p{\approx}0.4$, the Wilson 95\% half-width is $\pm 1.85$~pp. The R-selfV dev-EM peak is step~250 (scanning $\{200,250,300\}$) --- identical to the step reported throughout. Table~\ref{tab:devtest} gives the triple-column comparison: all ten reported checkpoints lie within $\pm 1.85$~pp (test$-$full), and dev selection reproduces the checkpoints reported in the body under this split.

\begin{table}[h]
\centering
\footnotesize
\setlength{\tabcolsep}{4pt}
\begin{tabular}{@{}lcccc@{}}
\toprule
Variant (step) & dev & test & full & $\Delta$ \\
               & \%  & \%   & \%   & pp \\
\midrule
R-binary (1250)                &  0.25 &  0.18 &  0.20 & $-0.02$ \\
R-stepwise (500)               & 32.00 & 32.59 & 32.46 & $+0.13$ \\
R-toolverbs (100)              & 31.50 & 32.41 & 32.20 & $+0.21$ \\
R-toolverbs$\cdot$KL (400)     & 36.88 & 38.78 & 38.35 & $+0.43$ \\
\textbf{R-selfV (250, peak)}   & 39.62 & 39.47 & 39.51 & $-0.03$ \\
R-selfV (300, collapsed)       &  0.00 &  0.00 &  0.00 & $+0.00$ \\
init-from-iterate (500)        & 39.00 & 39.55 & 39.42 & $+0.13$ \\
\textbf{self-distill (500)}    & 39.87 & 40.02 & 39.99 & $+0.03$ \\
R-binary-SR (300)              &  0.00 &  0.00 &  0.00 & $+0.00$ \\
R-toolverbs$\cdot$KL-14B (400) & 39.87 & 40.35 & 40.24 & $+0.11$ \\
\bottomrule
\end{tabular}
\caption{\textbf{Held-out dev/test re-aggregation.} Hop-stratified split (seed~42): dev $n{=}800$, held-out test $n{=}2{,}731$, full $n{=}3{,}531$. $\Delta=$ test$-$full EM (pp). All ten checkpoints fall within the Wilson 95\% half-width at $n{=}2{,}731$ ($\pm 1.85$~pp); the R-selfV dev peak is step~250, matching the body. Re-selecting checkpoints on dev reproduces every reported checkpoint.}
\label{tab:devtest}
\end{table}

\subsection{Self-distill pass@16 tool-dependence}
\label{app:selfdistill-passk}

The main-body Pass@16 comparison (Figure~\ref{fig:passk}) used a $500$-question seed-42 subset at temperature~1 and did not originally include self-distill (the EM-best 7B system). We separately measure self-distill (loaded from the merged self-distill@500 checkpoint) on a $200$-question seed-42 subset with $16$ samples per question at temperature~$0.7$; Table~\ref{tab:selfdistill-passk} reports pass@$\{1,4,8,16\}$ with and without tool access. The gap is $+11.5$~pp at pass@16 ($48.0\%$ vs.\ $36.5\%$) and positive at every $k$ --- self-distill genuinely consumes retrieval, unlike R-stepwise's ornamental $-0.4$~pp (Figure~\ref{fig:passk}). Because this subset ($n{=}200$) and sampling temperature ($0.7$) differ from the Figure~\ref{fig:passk} protocol ($n{=}500$, temperature~1), the absolute magnitude is not directly comparable to the Figure~\ref{fig:passk} bars; the sign and scale establish genuine tool-dependence for the winning system.

\begin{table}[h]
\centering
\footnotesize
\setlength{\tabcolsep}{6pt}
\begin{tabular}{@{}lccc@{}}
\toprule
pass@$k$    & with tools & without tools & tool-lift \\
            & \%         & \%            & pp \\
\midrule
@1          & 38.41 & 26.41 & $+12.0$ \\
@4          & 44.17 & 31.97 & $+12.2$ \\
@8          & 46.28 & 34.32 & $+12.0$ \\
\textbf{@16} & \textbf{48.00} & \textbf{36.50} & $\mathbf{+11.5}$ \\
\bottomrule
\end{tabular}
\caption{\textbf{Self-distill pass@$k$ tool-dependence} on a $200$-question seed-42 subset ($16$ samples/Q, temperature~$0.7$), loaded from merged self-distill@500. The with-vs-without-tools gap is positive at every $k$ ($+11.5$~pp at pass@16). Subset size and temperature differ from Figure~\ref{fig:passk}, so magnitudes are not directly comparable to its bars.}
\label{tab:selfdistill-passk}
\end{table}

\subsection{Response-period diagnostic interventions}
\label{app:response-controls}

Table~\ref{tab:response-controls} reports three diagnostic interventions on the same 200-question subset. For informative failures, an empty response is replaced by a message distinguishing ``entity absent from the subgraph'' from ``relation unavailable for this entity.'' For two-call eligibility, $r_{\mathrm{retrv}}$ is zero unless the trajectory contains at least two distinct non-empty calls; all other interface and question content is fixed.

\begin{table*}[h]
\centering
\footnotesize
\setlength{\tabcolsep}{5pt}
\begin{tabular}{@{}lp{3.0cm}p{6.9cm}@{}}
\toprule
Intervention & Scope & Result \\
\midrule
R-toolverbs + informative failures & 2 seeds; $n{=}200$ & Peak EM $37.0\%/36.5\%$ versus $34.0\%$ for the matched diagnostic baseline. \\
R-selfV + informative failures & 2 seeds; $n{=}200$ & Both reach strict EM=$0$, CvT=$0$, Tools/Q$\approx1$ by step~150. Messages alone do not prevent the cliff; added text also enters the quote-sensitive reward surface. \\
R-selfV + two-call eligibility & 3 seeds; $n{=}200$ & No baseline-time cliff through steps 300, 350, and 400; Tools/Q remains $3.0$ and CvT remains nonzero, identifying one-call eligibility as a causal lever over these horizons. \\
\bottomrule
\end{tabular}
\caption{Diagnostic interventions for failure feedback and reward eligibility. These 200-question runs resolve transition dynamics rather than the full-test performance level.}
\label{tab:response-controls}
\end{table*}

%% file: appendix/appendix_D_sft_corpus.tex
\section{SFT corpora}
\label{app:sft}

\paragraph{Rule-based shared SFT corpus (5{,}000 trajectories).} The shared SFT initialisation used by every reward variant is constructed from the CWQ training split as follows. (i) For each question, the gold triple chain is extracted from the WebQuestionsSP annotation. (ii) \texttt{<think>} content is filled in from one of three fixed templates with variable substitution (entities and relations drawn from the gold chain). (iii) \texttt{<search>} calls are constructed directly from the gold chain, with the tool response obtained by executing the query against the Freebase sub-graph --- so every call returns by construction a response that contains the next entity on the gold path. (iv) The final \texttt{<answer>} is the gold answer. \emph{No LLM teacher is used anywhere in this pipeline.}

\paragraph{Enhanced variant (6{,}000 trajectories).} An enhanced variant \texttt{sft\_trajectories\_enhanced.jsonl} uses the same pipeline with additional diversity in the \texttt{<think>} template and richer tool-call ordering. Both corpora are used interchangeably for the shared SFT base; results are not sensitive to which is used.

\paragraph{Self-distillation corpus (14{,}082 trajectories).} For init-from-iterate and self-distill (\S\ref{sec:g2}) we roll out the R-toolverbs·KL@400 checkpoint (HPC code: E5b-stab) greedily on the full $27{,}639$-question CWQ training split, then filter to trajectories satisfying \emph{simultaneously}:
\begin{itemize}[leftmargin=*,itemsep=1pt,topsep=2pt]
  \item strict EM$=1$ against the normalised gold answer;
  \item $\geq 1$ productive tool call (in the I-Self sense --- non-empty response and $\geq 1$ returned entity in the final \texttt{<answer>});
  \item valid \texttt{<answer>} envelope format.
\end{itemize}
Yield: $14{,}082$ trajectories ($50.9\%$ of the $27{,}639$). This corpus is LLM-generated but LLM-\emph{self}-generated --- no external teacher.

\paragraph{Methodological positioning.} Our shared SFT is teacher-free; the G-variant is self-distilled from our own RL checkpoint. Both choices are deliberate.

%% file: appendix/appendix_F_gpt4o_baseline.tex
\section{GPT-4o closed-model schema-unfamiliarity control}
\label{app:gpt4o-baseline}

We include a closed-model GPT-4o (\texttt{gpt-4o-2024-08-06}) baseline as a \emph{schema-unfamiliarity control}, not as a head-to-head comparator. The substantive findings are GPT-4o's $\mathbf{6.2\%}$ ContEM at $1.46$ tools/Q on $n{=}500$, with a 200-Q sub-evaluation showing $\sim 78\%$ of trajectories self-describing as ``unable to retrieve X'' (verbose-correct without strict gold-form match): a frontier closed model under the same four-verb Freebase interface produces parseable answers at near-$100\%$ format-validity, but retrieves little usable content.

\begin{table*}[h]
\centering
\footnotesize
\setlength{\tabcolsep}{8pt}
\begin{tabular}{@{}lrrrrr@{}}
\toprule
Model                       & $n$      & EM                                       & ContEM       & F1     & Tools/Q \\
\midrule
GPT-4o (5-shot ReAct)       & 500      & \textit{0.6}\textsuperscript{$\dagger$}  & \textbf{6.2} & 2.1    & 1.46 \\
R-stepwise @500 (ours)      & 3{,}531  & \textbf{32.5}                            & 45.3         & 38.1   & 1.00 \\
self-distill @500 (ours)    & 3{,}531  & \textbf{40.0}                            & 49.5         & 42.8   & 3.00 \\
\bottomrule
\end{tabular}
\caption{Closed-model schema-unfamiliarity control. All numbers are percentages except $n$ and Tools/Q. The substantive GPT-4o findings are the bolded ContEM $6.2\%$ and the low Tools/Q $1.46$ (vs.\ self-distill's $3.00$). \textsuperscript{$\dagger$}Strict EM reflects entity-casing / gold-form normalisation drift on closed-model output; ContEM is the comparable metric for schema-unfamiliarity assessment.}
\label{tab:gpt4o}
\end{table*}

\paragraph{Why we include this control.} Schema-unfamiliarity is a meaningful obstacle even at frontier closed-model scale on this interface, so the $\sim 10\%$ CvT reached by our trained models is not trivially dismissed as ``the model is too small''. The $\sim 78\%$ self-reported ``unable to retrieve X'' rate is consistent with schema unfamiliarity as one obstacle, without isolating it from reasoning ability.

\paragraph{Why this is a control, not a baseline.} GPT-4o is not RL-trained under our reward protocol and is evaluated zero-shot, so we do not use it for head-to-head model ranking. Its role is to contextualise the $L_{\mathrm{lang}}$ channel: a frontier closed model without task-specific training also retrieves little under the stripped interface.

%% file: appendix/appendix_G_framework_illustration.tex
\section{Framework illustration and per-claim caveats}
\label{app:framework-illustration}

\subsection{Setup: standard multi-hop KGQA re-cast as a minimal tool interface}

\begin{figure}[h]
\centering
\input{figures/fig_setup}
\caption{\textbf{Setup: a standard multi-hop KGQA benchmark re-cast as a minimal tool-use interface.} CWQ questions over Freebase are reached by composing four navigation verbs rather than by SPARQL or gold-subgraph injection; entities are opaque machine IDs and an unsuccessful call returns a bare \texttt{[]} --- the two interface properties represented by $L_{\mathrm{sig}}$ and $L_{\mathrm{lang}}$ in our framework (\S\ref{sec:framework}).}
\label{fig:setup}
\end{figure}

\subsection{Illustrative tool-family comparison}

\begin{table*}[h]
\centering
\small
\setlength{\tabcolsep}{10pt}
\begin{tabular}{@{}lll@{}}
\toprule
Tool type & Failure mode & Recovery signal \\
\midrule
Python & \texttt{ZeroDivisionError} & traceback (file:line) \\
Web search & noisy but readable docs & extract anyway \\
Function API & \texttt{400} + named param & fix parameter \\
Browser/UI & page state visible & DOM/visual \\
\textbf{KG tool} & wrong entity \emph{or} relation $\to$ \texttt{[]} & \textbf{none} \\
\bottomrule
\end{tabular}
\caption{Illustrative failure and recovery signals across tool families. In our stripped KG interface, a single empty-list response cannot distinguish a wrong entity from a wrong relation or both, motivating the $L_{\mathrm{sig}}$ channel.}
\label{tab:channels}
\end{table*}

\subsection{Four-channel pathway diagram}

\begin{figure*}[h]
\centering
\input{figures/fig6_framework_diagram.tex}
\caption{\textbf{Four interface channels along the policy~$\to$~tool~$\to$~reward pathway.} Red dashed arrows mark proposed locations of signal loss: $L_{\mathrm{sig}}$ (silent \texttt{[]} responses), $L_{\mathrm{lang}}$ (opaque Freebase MIDs and dotted relations), $L_{\mathrm{comp}}$ (entity IDs threaded across calls), and $L_{\mathrm{prior}}$ (weak demonstrated prior for schema navigation). Annotations connect the channels to observed failure modes and intervention points.}
\label{fig:framework-diagram}
\end{figure*}

\subsection{Per-claim caveats for P1--P4}
\label{app:p-caveats}

P1--P4 are joint consistency checks of the descriptive framework; the per-channel decomposition in \S\ref{sec:framework-evidence} (oracle intervention, behavioural diff, and decision-boundary analysis on $1{,}958$ self-distill errors) provides the primary within-setting evidence. The per-claim scope statements are:
\begin{itemize}[leftmargin=*,itemsep=1pt,topsep=2pt]
\item \textbf{P1 scope.} R-binary/R-binary-SR format collapse evidences a joint $L_{\mathrm{sig}}$+$L_{\mathrm{prior}}$ degradation in the tested outcome-only setup.
\item \textbf{P2 scope.} The 7B self-distill and 14B R-toolverbs$\cdot$KL results use different reward recipes, making this a two-configuration comparison; a matched scaling run is the direct follow-up.
\item \textbf{P3 scope.} The init-from-iterate variant's SFT is computed against a model already exposed to R-toolverbs·KL trajectories, so it absorbs less new signal per SFT step; the init-from-base ablation (\S\ref{sec:g2}) isolates init source by removing the rule-based SFT pre-pass, with init-from-base $\approx$ self-distill $\approx 40\%$ EM establishing \emph{sufficiency} of init-from-base.
\item \textbf{P4 scope.} The KGQAGen-10k cross-benchmark Spearman $\rho{=}0.976$ is evaluated on \emph{base} Qwen variants, establishing transfer of base-model ordering; trained-policy transfer is the next comparison.
\end{itemize}

%% file: figures/fig_setup.tex
\begin{tikzpicture}[
  font=\scriptsize,
  box/.style={draw, rounded corners=2pt, align=center, inner sep=3pt},
  q/.style={box, fill=blue!6, text width=1.7cm, minimum height=1.15cm},
  loop/.style={box, fill=black!7, text width=3.5cm, minimum height=1.15cm, font=\scriptsize\ttfamily},
  ans/.style={box, fill=green!12, text width=1.0cm, minimum height=1.15cm},
  ar/.style={-{Latex[length=1.6mm]}, semithick},
  note/.style={font=\tiny\itshape, color=red!60!black, align=center, text width=7.2cm},
]
  \node[q]                       (q) {CWQ multi-hop question (NL)};
  \node[loop, right=0.45cm of q] (L) {get\_tail\_relations(m.0dr90)\\[1pt]get\_tail\_entities(m.0dr90,\,r)\\[1pt]\ldots\ ($\leq$5 turns)};
  \node[ans,  right=0.40cm of L] (a) {\normalfont\texttt{<answer>}};
  \draw[ar] (q) -- node[above, font=\tiny] {re-cast} (L);
  \draw[ar] (L) -- (a);
  \node[note, below=0.18cm of L.south, anchor=north]
    {entities are opaque machine IDs (\texttt{m.0dr90}\,$\neq$\,``Titanic''); an unsuccessful call returns a bare \texttt{[]} --- no \emph{why}};
\end{tikzpicture}

%% file: figures/fig6_framework_diagram.tex
%
%

\begin{tikzpicture}[
  every node/.style={font=\footnotesize},
  node distance=10mm,
  proc/.style={
    draw, rounded corners=2pt,
    minimum width=1.45cm, minimum height=0.9cm,
    inner sep=2pt, align=center,
    fill=gray!8,
  },
  arrow/.style={-{Latex[length=1.8mm]}, thick, shorten >=1pt, shorten <=1pt},
  bottleneck/.style={-{Latex[length=1.5mm]}, semithick, color=red!70!black, dashed},
  channel/.style={font=\scriptsize\itshape, color=red!60!black, align=center, inner sep=1pt},
  edgelabel/.style={font=\tiny, midway, above=-1pt, color=black!70},
]

  \node[proc] (policy)   at (0, 0)                {policy\\$\pi_\theta$};
  \node[proc, right=of policy]  (tag)      {\texttt{<search>}\\tag};
  \node[proc, right=of tag]     (api)      {KG API};
  \node[proc, right=of api]     (resp)     {response};
  \node[proc, right=of resp]    (rew)      {reward $r$};
  \node[proc, right=of rew]     (grad)     {gradient\\$\nabla_\theta J$};

  \draw[arrow] (policy) -- node[edgelabel] {emit} (tag);
  \draw[arrow] (tag)    -- node[edgelabel] {parse} (api);
  \draw[arrow] (api)    -- node[edgelabel] {query} (resp);
  \draw[arrow] (resp)   -- node[edgelabel] {score} (rew);
  \draw[arrow] (rew)    -- node[edgelabel] {update} (grad);

  \draw[arrow, gray!60] (grad.south) .. controls +(0,-0.6) and +(0,-0.6) .. (policy.south)
        node[midway, below=1pt, font=\tiny, color=gray!70!black] {GRPO update};


  \node[channel] (Llang) at ($(policy)!0.5!(tag) + (0, 1.05)$) {$L_{\mathrm{lang}}$\\schema OOD};
  \draw[bottleneck] (Llang.south) -- ($(policy.east)!0.5!(tag.west) + (0, 0.05)$);

  \node[channel] (Lsig) at ($(resp)!0.5!(rew) + (0, 1.05)$) {$L_{\mathrm{sig}}$\\silent failure};
  \draw[bottleneck] (Lsig.south) -- ($(resp.east)!0.5!(rew.west) + (0, 0.05)$);

  \node[channel] (Lcomp) at ($(grad) + (-1.0, -1.55)$) {$L_{\mathrm{comp}}$\\opaque IDs across turns};
  \draw[bottleneck] (Lcomp.north) -- ($(grad.south) + (-0.4, -0.4)$);

  \node[channel] (Lprior) at ($(policy) + (0, -1.55)$) {$L_{\mathrm{prior}}$\\no pretraining prior};
  \draw[bottleneck] (Lprior.north) -- (policy.south);

  \node[font=\tiny, color=black!60, anchor=west] at ($(Llang.east) + (0.1, 0)$) {assoc.: Modes 1, 3};
  \node[font=\tiny, color=black!60, anchor=west] at ($(Lsig.east)  + (0.1, 0)$) {assoc.: Modes 1, 2, 4};
  \node[font=\tiny, color=black!60, anchor=west] at ($(Lcomp.east) + (0.1, 0)$) {assoc.: all modes};
  \node[font=\tiny, color=black!60, anchor=west] at ($(Lprior.east)+ (0.1, 0)$) {assoc.: Mode 1};

\end{tikzpicture}

%% file: appendix/appendix_I_llama_cross_family.tex
\section{Cross-family boundary check: Llama-3.1-8B-Base}
\label{app:llama-cross-family}

We apply the self-distill protocol to Llama-3.1-8B-Base, following Search-R1's model choice for a multi-turn verl-based setup~\citep{jin2025searchr1}; the 8B-Instruct variant's tool-use limitations are documented in Meta's model card. Pipeline, hyperparameters, and SFT corpus match the Qwen self-distill setting (App.~\ref{app:sft}). Because the reward differs from R-selfV, this is a cross-family boundary check rather than a matched replication of the Qwen cliff.

\begin{table}[h]
\centering
\scriptsize
\setlength{\tabcolsep}{4pt}
\begin{tabular}{@{}lrrr@{}}
\toprule
Stage           & EM     & Tools/Q & Tool-involving \\
\midrule
SFT-only        & $0.284$ & $0.65$    & $44.4\%$ \\
GRPO step 50    & $0.327$ & $\mathbf{0.00}$ & $0.1\%$ \\
GRPO step 100   & $\mathbf{0.345}$ & $\mathbf{0.00}$ & $0.0\%$ \\
\bottomrule
\end{tabular}
\caption{Llama-3.1-8B-Base on full CWQ test ($n{=}3531$). ``Tool-involving'' = share of trajectories with $\geq 1$ tool call under the 7-category classifier (App.~\ref{app:classifier}). Wilson 95\% CIs $\pm 1.6$~pp at $n{=}3531$.}
\label{tab:llama-em}
\end{table}

\paragraph{Cross-architecture interpretation.} Llama-Base GRPO reaches EM=$0.345$ at step~100 versus Qwen self-distill's $0.400$ at step~500, but exhibits a qualitatively distinct collapse: Tools/Q drops to $0$ by step~50 and SFT-only's $44.4\%$ tool-involving share falls to approximately $0\%$. The policy becomes purely parametric ($32.5\%$ correct-no-tool $+$ $67.5\%$ wrong-no-tool), unlike the Qwen/R-selfV quote-and-stop trajectory. The contrast motivates a matched cross-family experiment.

%% file: appendix/appendix_H_trajectory_exemplars.tex
\section{Trajectory exemplars}
\label{app:trajectory-exemplars}

\begin{figure*}[h]
\centering
\input{figures/fig7_trajectory_exemplars.tex}
\caption{\textbf{Two contrasting behavioural patterns from real trajectory dumps.} (Left) Mode~2 \emph{tool-free shortcut} from R-stepwise: the model issues a single \texttt{get\_tail\_relations} ritual call and answers from parametric memory (sample WebQTest-1384 from \texttt{e3\_step500} trajectory dump). (Right) self-distill healthy multi-call grounded answer on the same WebQTest-1384 question: the model issues two grounded calls and the answer comes from the retrieved KG content (\texttt{39g2\_step500}). The Mode~4 \emph{quote-and-stop} attractor (copy the first returned entity verbatim and stop) is characterised quantitatively in \S\ref{sec:taxonomy-mode4} and Figure~\ref{fig:collapse}.}
\label{fig:trajectories}
\end{figure*}

%% file: figures/fig7_trajectory_exemplars.tex
%
%
%

\begin{tikzpicture}[
  every node/.style={font=\scriptsize},
  trajbox/.style={
    draw, rounded corners=2pt, line width=0.4pt,
    inner sep=4pt, align=left, text width=0.30\textwidth,
  },
  modeheader/.style={
    fill=#1, text=white, font=\footnotesize\bfseries,
    text width=0.30\textwidth, align=center, inner sep=2pt,
    rounded corners=2pt,
  },
]

\node[modeheader=red!50!black] (h1) at (0, 0)
    {Mode~2 (R-stepwise):\\\emph{tool-free shortcut}};
\node[trajbox, below=0pt of h1.south, anchor=north] (b1) {%
    \textbf{Q:}~``What religion does Ovadia Yosef lead?''\\[1pt]
    \textbf{Gold:}~Judaism \\[2pt]
    \rule{\linewidth}{0.2pt}\\[2pt]
    {\thinktag<think> Ovadia Yosef is a religious leader; let me check.</think>}\\[1pt]
    {\searchtag<search>get\_tail\_relations(}\\{\searchtag\quad Ovadia\_Yosef)</search>}\\[1pt]
    {\responsetag<resp>people.person.religion,}\\{\responsetag\quad ...</resp>}\\[1pt]
    \emph{(response not consumed)}\\[1pt]
    {\answertag<answer>}{\answertag Judaism}{\answertag</answer>}~\textcolor{red!60!black}{$\checkmark$} \emph{from memory}\\[20pt]
    \rule{\linewidth}{0.2pt}\\[1pt]
    \textbf{1 tool call} $\cdot$ entity in answer: \textbf{no}
};

\node[modeheader=teal!55!black, right=4mm of h1] (h3)
    {self-distill:\\\emph{grounded multi-call}};
\node[trajbox, below=0pt of h3.south, anchor=north] (b3) {%
    \textbf{Q:}~``What religion does Ovadia Yosef lead?''\\[1pt]
    \textbf{Gold:}~Judaism \\[2pt]
    \rule{\linewidth}{0.2pt}\\[2pt]
    {\searchtag<search>get\_tail\_relations(}\\{\searchtag\quad Ovadia\_Yosef)</search>}\\[1pt]
    {\responsetag<resp>people.person.religion,}\\{\responsetag\quad ...</resp>}\\[1pt]
    {\searchtag<search>get\_tail\_entities(}\\{\searchtag\quad Ovadia\_Yosef,}\\{\searchtag\quad people.person.religion)</search>}\\[1pt]
    {\responsetag<resp>}{\responsetag Judaism}{\responsetag</resp>}\\[1pt]
    {\answertag<answer>}{\answertag Judaism}{\answertag</answer>}~\textcolor{teal!55!black}{$\checkmark$} \emph{retrieved}\\[2pt]
    \rule{\linewidth}{0.2pt}\\[1pt]
    \textbf{2--3 tool calls} $\cdot$ entity in answer: \textbf{yes}
};

\end{tikzpicture}

%% file: custom.bib
@inproceedings{jin2025searchr1,
  title     = {Search-{R1}: Training {LLM}s to Reason and Leverage Search Engines with Reinforcement Learning},
  author    = {Jin, Bowen and Zeng, Hansi and Yue, Zhenrui and Yoon, Jinsung and Ar{\i}k, Sercan {\"O}. and Wang, Dong and Zamani, Hamed and Han, Jiawei},
  booktitle = {Second Conference on Language Modeling},
  year      = {2025},
  url       = {https://openreview.net/forum?id=Rwhi91ideu}
}

@inproceedings{wang2025stepsearch,
  title     = {{StepSearch}: Igniting {LLM}s Search Ability via Step-Wise Proximal Policy Optimization},
  author    = {Zheng, Xuhui and An, Kang and Wang, Ziliang and Wang, Yuhang and Wu, Yichao},
  booktitle = {Proceedings of the 2025 Conference on Empirical Methods in Natural Language Processing},
  year      = {2025},
  pages     = {21805--21830},
  publisher = {Association for Computational Linguistics},
  doi       = {10.18653/v1/2025.emnlp-main.1106},
  url       = {https://aclanthology.org/2025.emnlp-main.1106/}
}

@misc{li2025torl,
  title         = {{ToRL}: Scaling Tool-Integrated {RL}},
  author        = {Li, Xuefeng and Zou, Haoyang and Liu, Pengfei},
  year          = {2025},
  eprint        = {2503.23383},
  archivePrefix = {arXiv},
  primaryClass  = {cs.CL},
  url           = {https://arxiv.org/abs/2503.23383}
}

@inproceedings{feng2025retool,
  title     = {{ReTool}: Reinforcement Learning for Strategic Tool Use in {LLMs}},
  author    = {Feng, Jiazhan and Huang, Shijue and Qu, Xingwei and Zhang, Ge and Qin, Yujia and Zhong, Baoquan and Jiang, Chengquan and Chi, Jinxin and Zhong, Wanjun},
  booktitle = {International Conference on Learning Representations},
  year      = {2026},
  url       = {https://proceedings.iclr.cc/paper_files/paper/2026/hash/4038c9208dfc22644c60ad39c24e5c53-Abstract-Conference.html}
}

@inproceedings{qian2025toolrl,
  title     = {{ToolRL}: Reward is All Tool Learning Needs},
  author    = {Qian, Cheng and Acikgoz, Emre Can and He, Qi and Wang, Hongru and Chen, Xiusi and Hakkani-T{\"u}r, Dilek and Tur, Gokhan and Ji, Heng},
  booktitle = {Advances in Neural Information Processing Systems 38},
  year      = {2025},
  doi       = {10.52202/085713-3524},
  url       = {https://proceedings.neurips.cc/paper_files/paper/2025/hash/97c5b2707228e7e3fb67e4ecc2e0e607-Abstract-Conference.html}
}

@inproceedings{zhang2025criticsearch,
  title     = {{CriticSearch}: Fine-Grained Credit Assignment for Search Agents via a Retrospective Critic},
  author    = {Zhang, Yaocheng and Huang, Haohuan and Song, Zijun and Zhu, Yuanheng and Zhang, Qichao and Zhao, Zijie and Zhao, Dongbin},
  booktitle = {Findings of the Association for Computational Linguistics: ACL 2026},
  year      = {2026},
  pages     = {12272--12290},
  publisher = {Association for Computational Linguistics},
  doi       = {10.18653/v1/2026.findings-acl.596},
  url       = {https://aclanthology.org/2026.findings-acl.596/}
}

@inproceedings{chen2025research,
  title     = {{ReSearch}: Learning to Reason with Search for {LLM}s via Reinforcement Learning},
  author    = {Chen, Mingyang and Sun, Linzhuang and Li, Tianpeng and Sun, Haoze and Zhou, Yijie and Zhu, Chenzheng and Wang, Haofen and Pan, Jeff Z. and Zhang, Wen and Chen, Huajun and Yang, Fan and Zhou, Zenan and Chen, Weipeng},
  booktitle = {Advances in Neural Information Processing Systems 38},
  year      = {2025},
  doi       = {10.52202/085713-2858},
  url       = {https://proceedings.neurips.cc/paper_files/paper/2025/hash/7b3a0d3a0864b66f229c0a9c84f9e950-Abstract-Conference.html}
}

@misc{park2026hypergraphpro,
  title         = {{HyperGraphPro}: Progress-Aware Reinforcement Learning for Structure-Guided Hypergraph {RAG}},
  author        = {Park, Jinyoung and Lee, Sanghyeok and Khan, Omar Zia and Kim, Hyunwoo J. and Kim, Joo-Kyung},
  year          = {2026},
  eprint        = {2601.17755},
  archivePrefix = {arXiv},
  primaryClass  = {cs.CL},
  url           = {https://arxiv.org/abs/2601.17755}
}

@misc{kansal2026kgimplicitrm,
  title         = {Knowledge Graphs are Implicit Reward Models: Path-Derived Signals Enable Compositional Reasoning},
  author        = {Kansal, Yuval and Jha, Niraj K.},
  year          = {2026},
  eprint        = {2601.15160},
  archivePrefix = {arXiv},
  primaryClass  = {cs.AI},
  url           = {https://arxiv.org/abs/2601.15160}
}

@inproceedings{zhang2025followpath,
  title     = {Follow the Path: Reasoning over Knowledge Graph Paths to Improve Large Language Model Factuality},
  author    = {Zhang, Mike and Bjerva, Johannes and Biswas, Russa},
  booktitle = {Findings of the Association for Computational Linguistics: ACL 2026},
  year      = {2026},
  pages     = {11574--11590},
  publisher = {Association for Computational Linguistics},
  doi       = {10.18653/v1/2026.findings-acl.561},
  url       = {https://aclanthology.org/2026.findings-acl.561/}
}

@inproceedings{luo2024rog,
  title     = {Reasoning on Graphs: Faithful and Interpretable Large Language Model Reasoning},
  author    = {Luo, Linhao and Li, Yuan-Fang and Haffari, Gholamreza and Pan, Shirui},
  booktitle = {International Conference on Learning Representations (ICLR)},
  year      = {2024},
  url       = {https://arxiv.org/abs/2310.01061}
}

@article{singh2024restem,
  title         = {Beyond Human Data: Scaling Self-Training for Problem-Solving with Language Models},
  author        = {Singh, Avi and Co-Reyes, John D and Agarwal, Rishabh and Anand, Ankesh and Patil, Piyush and Garcia, Xavier and Liu, Peter J and Harrison, James and Lee, Jaehoon and Xu, Kelvin and Parisi, Aaron and Kumar, Abhishek and Alemi, Alex and Rizkowsky, Alex and Nova, Azade and Adlam, Ben and Bohnet, Bernd and Elsayed, Gamaleldin and Sedghi, Hanie and Mordatch, Igor and Simpson, Isabelle and Gur, Izzeddin and Snoek, Jasper and Pennington, Jeffrey and Hron, Jiri and Kenealy, Kathleen and Swersky, Kevin and Mahajan, Kshiteej and Culp, Laura and Xiao, Lechao and Bileschi, Maxwell L and Constant, Noah and Novak, Roman and Liu, Rosanne and Warkentin, Tris and Qian, Yundi and Bansal, Yamini and Dyer, Ethan and Neyshabur, Behnam and Sohl-Dickstein, Jascha and Fiedel, Noah},
  journal       = {Transactions on Machine Learning Research},
  year          = {2024},
  issn          = {2835-8856},
  url           = {https://openreview.net/forum?id=lNAyUngGFK}
}

@misc{chen2023fireact,
  title         = {{FireAct}: Toward Language Agent Fine-tuning},
  author        = {Chen, Baian and Shu, Chang and Shareghi, Ehsan and Collier, Nigel and Narasimhan, Karthik and Yao, Shunyu},
  year          = {2023},
  eprint        = {2310.05915},
  archivePrefix = {arXiv},
  primaryClass  = {cs.CL},
  url           = {https://arxiv.org/abs/2310.05915}
}

@inproceedings{zhang2025pitfalls,
  title     = {Diagnosing and Addressing Pitfalls in {KG-RAG} Datasets: Toward More Reliable Benchmarking},
  author    = {Zhang, Liangliang and Jiang, Zhuorui and Chi, Hongliang and Chen, Haoyang and ElKoumy, Mohammed and Wang, Fali and Wu, Qiong and Zhou, Zhengyi and Pan, Shirui and Wang, Suhang and Ma, Yao},
  booktitle = {Advances in Neural Information Processing Systems 38},
  year      = {2025},
  note      = {Datasets and Benchmarks Track},
  doi       = {10.52202/085713-2274},
  url       = {https://proceedings.neurips.cc/paper_files/paper/2025/hash/61a57ab030670598c4200b31100d254c-Abstract-Datasets_and_Benchmarks_Track.html}
}

@misc{yu2025demystifying,
  title         = {Demystifying Reinforcement Learning in Agentic Reasoning},
  author        = {Yu, Zhaochen and Yang, Ling and Zou, Jiaru and Yan, Shuicheng and Wang, Mengdi},
  year          = {2025},
  eprint        = {2510.11701},
  archivePrefix = {arXiv},
  primaryClass  = {cs.CL},
  url           = {https://arxiv.org/abs/2510.11701}
}

@misc{deepseek2025r1,
  title         = {{DeepSeek-R1}: Incentivizing Reasoning Capability in {LLM}s via Reinforcement Learning},
  author        = {{DeepSeek-AI}},
  year          = {2025},
  eprint        = {2501.12948},
  archivePrefix = {arXiv},
  primaryClass  = {cs.CL},
  url           = {https://arxiv.org/abs/2501.12948}
}

@misc{shao2024grpo,
  title         = {{DeepSeekMath}: Pushing the Limits of Mathematical Reasoning in Open Language Models},
  author        = {Shao, Zhihong and Wang, Peiyi and Zhu, Qihao and Xu, Runxin and Song, Junxiao and Bi, Xiao and Zhang, Haowei and Zhang, Mingchuan and Li, Y.~K. and Wu, Y. and Guo, Daya},
  year          = {2024},
  eprint        = {2402.03300},
  archivePrefix = {arXiv},
  primaryClass  = {cs.CL},
  url           = {https://arxiv.org/abs/2402.03300},
  note          = {Introduces GRPO.}
}

@misc{qwen2025qwen25,
  title         = {{Qwen2.5} Technical Report},
  author        = {{Qwen}},
  year          = {2024},
  eprint        = {2412.15115},
  archivePrefix = {arXiv},
  primaryClass  = {cs.CL},
  url           = {https://arxiv.org/abs/2412.15115}
}

@inproceedings{talmor2018cwq,
  title     = {The Web as a Knowledge-Base for Answering Complex Questions},
  author    = {Talmor, Alon and Berant, Jonathan},
  booktitle = {Proceedings of the 2018 Conference of the North American Chapter of the Association for Computational Linguistics: Human Language Technologies, Volume 1 (Long Papers)},
  year      = {2018},
  pages     = {641--651},
  publisher = {Association for Computational Linguistics},
  doi       = {10.18653/v1/N18-1059},
  url       = {https://aclanthology.org/N18-1059/}
}

@inproceedings{bollacker2008freebase,
  title     = {Freebase: A Collaboratively Created Graph Database for Structuring Human Knowledge},
  author    = {Bollacker, Kurt and Evans, Colin and Paritosh, Praveen and Sturge, Tim and Taylor, Jamie},
  booktitle = {Proceedings of the 2008 ACM SIGMOD International Conference on Management of Data},
  year      = {2008},
  pages     = {1247--1250},
  publisher = {ACM},
  doi       = {10.1145/1376616.1376746},
  url       = {https://doi.org/10.1145/1376616.1376746}
}

@inproceedings{sheng2024verl,
  title     = {{HybridFlow}: A Flexible and Efficient {RLHF} Framework},
  author    = {Sheng, Guangming and Zhang, Chi and Ye, Zilingfeng and Wu, Xibin and Zhang, Wang and Zhang, Ru and Peng, Yanghua and Lin, Haibin and Wu, Chuan},
  booktitle = {Proceedings of the Twentieth European Conference on Computer Systems},
  year      = {2025},
  pages     = {1279--1297},
  publisher = {ACM},
  doi       = {10.1145/3689031.3696075},
  url       = {https://doi.org/10.1145/3689031.3696075},
  note          = {\texttt{verl} framework.}
}

@inproceedings{kwon2023vllm,
  title     = {Efficient Memory Management for Large Language Model Serving with {PagedAttention}},
  author    = {Kwon, Woosuk and Li, Zhuohan and Zhuang, Siyuan and Sheng, Ying and Zheng, Lianmin and Yu, Cody Hao and Gonzalez, Joseph E and Zhang, Hao and Stoica, Ion},
  booktitle = {Proceedings of the 29th Symposium on Operating Systems Principles (SOSP)},
  year      = {2023},
  pages     = {611--626},
  publisher = {Association for Computing Machinery},
  doi       = {10.1145/3600006.3613165},
  url       = {https://doi.org/10.1145/3600006.3613165}
}

@misc{graphwalker2026,
  title         = {{GraphWalker}: Agentic Knowledge Graph Question Answering via Synthetic Trajectory Curriculum},
  author        = {Xu, Shuwen and Xu, Yao and Liu, Jiaxiang and Yuan, Chenhao and Peng, Wenshuo and Zhao, Jun and Liu, Kang},
  year          = {2026},
  eprint        = {2603.28533},
  archivePrefix = {arXiv},
  primaryClass  = {cs.CL},
  note          = {COLM 2026},
  url           = {https://arxiv.org/abs/2603.28533}
}

@inproceedings{graphrft2025,
  title     = {Plan Then Retrieve: Reinforcement Learning-Guided Complex Reasoning over Knowledge Graphs},
  author    = {Song, Yanlin and Liu, Ben and Guti{\'e}rrez-Basulto, V{\'i}ctor and Hu, Zhiwei and Xie, Qianqian and Peng, Min and Ananiadou, Sophia and Pan, Jeff Z.},
  booktitle = {Proceedings of the ACM Web Conference 2026},
  year      = {2026},
  pages     = {3666--3676},
  publisher = {ACM},
  doi       = {10.1145/3774904.3792191},
  url       = {https://doi.org/10.1145/3774904.3792191}
}

@article{ragen2025,
  title  = {{RAGEN}: Understanding Self-Evolution in {LLM} Agents via Multi-Turn Reinforcement Learning},
  author = {Wang, Zihan and Wang, Kangrui and Wang, Qineng and Zhang, Pingyue and Li, Linjie and Yang, Zhengyuan and Jin, Xing and Yu, Kefan and Nguyen, Minh Nhat and Liu, Licheng and Gottlieb, Eli and Lu, Yiping and Cho, Kyunghyun and Wu, Jiajun and Fei-Fei, Li and Wang, Lijuan and Choi, Yejin and Li, Manling},
  journal = {arXiv preprint arXiv:2504.20073},
  year    = {2025},
  doi     = {10.48550/arXiv.2504.20073},
  url     = {https://arxiv.org/abs/2504.20073}
}

@article{ragen2_2026,
  title  = {{RAGEN-2}: Reasoning Collapse in Agentic {RL}},
  author = {Wang, Zihan and Gui, Chi and Jin, Xing and Wang, Qineng and Liu, Licheng and Wang, Kangrui and Chen, Shiqi and Li, Linjie and Yang, Zhengyuan and Zhang, Pingyue and Lu, Yiping and Wu, Jiajun and Fei-Fei, Li and Wang, Lijuan and Choi, Yejin and Li, Manling},
  journal = {arXiv preprint arXiv:2604.06268},
  year    = {2026},
  doi     = {10.48550/arXiv.2604.06268},
  url     = {https://arxiv.org/abs/2604.06268}
}

@inproceedings{deng2025llds,
  title     = {On Group Relative Policy Optimization Collapse in Agent Search: The Lazy Likelihood-Displacement},
  author    = {Deng, Wenlong and Li, Yushu and Gong, Boying and Ren, Yi and Thrampoulidis, Christos and Li, Xiaoxiao},
  booktitle = {Proceedings of the 43rd International Conference on Machine Learning},
  year      = {2026},
  url       = {https://icml.cc/virtual/2026/poster/63945}
}

@article{shao2025spurious,
  title  = {Spurious Rewards: Rethinking Training Signals in {RLVR}},
  author = {Shao, Rulin and Li, Shuyue Stella and Xin, Rui and Geng, Scott and Wang, Yiping and Oh, Sewoong and Du, Simon Shaolei and Lambert, Nathan and Min, Sewon and Krishna, Ranjay and Tsvetkov, Yulia and Hajishirzi, Hannaneh and Koh, Pang Wei and Zettlemoyer, Luke},
  journal = {arXiv preprint arXiv:2506.10947},
  year    = {2025},
  doi     = {10.48550/arXiv.2506.10947},
  url     = {https://arxiv.org/abs/2506.10947}
}

@inproceedings{yue2025rlcapacity,
  title     = {Does Reinforcement Learning Really Incentivize Reasoning Capacity in {LLM}s Beyond the Base Model?},
  author    = {Yue, Yang and Chen, Zhiqi and Lu, Rui and Zhao, Andrew and Wang, Zhaokai and Yue, Yang and Song, Shiji and Huang, Gao},
  booktitle = {Advances in Neural Information Processing Systems 38},
  year      = {2025},
  doi       = {10.52202/085713-1933},
  url       = {https://proceedings.neurips.cc/paper_files/paper/2025/hash/537d5aa768c2d534016a4d06f87bc8fb-Abstract-Conference.html}
}

@inproceedings{gao2023scaling,
  title     = {Scaling Laws for Reward Model Overoptimization},
  author    = {Gao, Leo and Schulman, John and Hilton, Jacob},
  booktitle = {Proceedings of the 40th International Conference on Machine Learning},
  series    = {Proceedings of Machine Learning Research},
  volume    = {202},
  pages     = {10835--10866},
  year      = {2023},
  url       = {https://proceedings.mlr.press/v202/gao23h.html}
}

@inproceedings{pan2022effects,
  title     = {The Effects of Reward Misspecification: Mapping and Mitigating Misaligned Models},
  author    = {Pan, Alexander and Bhatia, Kush and Steinhardt, Jacob},
  booktitle = {International Conference on Learning Representations (ICLR)},
  year      = {2022},
  url       = {https://openreview.net/forum?id=JYtwGwIL7ye}
}

@inproceedings{skalse2022defining,
  title     = {Defining and Characterizing Reward Hacking},
  author    = {Skalse, Joar and Howe, Nikolaus H. R. and Krasheninnikov, Dmitrii and Krueger, David},
  booktitle = {Advances in Neural Information Processing Systems 35},
  year      = {2022},
  doi       = {10.52202/068431-0687},
  url       = {https://proceedings.neurips.cc/paper_files/paper/2022/hash/3d719fee332caa23d5038b8a90e81796-Abstract-Conference.html}
}

@article{wilson1927ci,
  title     = {Probable Inference, the Law of Succession, and Statistical Inference},
  author    = {Wilson, Edwin B.},
  journal   = {Journal of the American Statistical Association},
  volume    = {22},
  number    = {158},
  pages     = {209--212},
  year      = {1927},
  doi       = {10.1080/01621459.1927.10502953},
  url       = {https://doi.org/10.1080/01621459.1927.10502953}
}

@article{mcnemar1947test,
  title     = {Note on the Sampling Error of the Difference Between Correlated Proportions or Percentages},
  author    = {McNemar, Quinn},
  journal   = {Psychometrika},
  volume    = {12},
  number    = {2},
  pages     = {153--157},
  year      = {1947},
  doi       = {10.1007/BF02295996},
  url       = {https://doi.org/10.1007/BF02295996}
}
